\documentclass[10pt,twocolumn,letterpaper]{article}

\usepackage{wacv}
\usepackage{times}
\usepackage{epsfig}
\usepackage{graphicx}
\usepackage{amsmath}
\usepackage{amssymb}
\usepackage{verbatim}
\usepackage{makecell}

\usepackage{booktabs}
\usepackage{multirow}
\usepackage{bm}
\DeclareMathAlphabet\mathbfcal{OMS}{cmsy}{b}{n}

\usepackage[table,xcdraw]{xcolor}
\usepackage{subcaption}

\usepackage{pdfpages}

\def\wacvPaperID{774} 

\wacvfinalcopy 
\ifwacvfinal
\def\assignedStartPage{9876} 
\fi

\ifwacvfinal
\usepackage[breaklinks=true,bookmarks=false]{hyperref}
\else
\usepackage[pagebackref=true,breaklinks=true,colorlinks,bookmarks=false]{hyperref}
\fi

\ifwacvfinal
\else
\fi

\renewcommand\footnotemark{}

\newcommand{\revised}[1]{\textcolor{black}{#1}}

\begin{document}

\title{Unsupervised Domain Adaptation in Semantic Segmentation \\via Orthogonal and Clustered Embeddings 
}

\author{Marco Toldo\thanks{Our work was in part supported by the Italian Minister for Education (MIUR) under the ``Departments of Excellence" initiative (Law 232/2016).}, 
Umberto Michieli, Pietro Zanuttigh\\
Department of Information Engineering, University of Padova\\
{\tt\small \{toldomarco,umberto.michieli,zanuttigh\}@dei.unipd.it}
}

\maketitle

\begin{abstract}

Deep learning frameworks allowed for a remarkable advancement in semantic segmentation, 
but the data hungry nature of convolutional networks has rapidly raised 
the demand for adaptation techniques  able to  transfer learned knowledge from  label-abundant domains  to  unlabeled ones. 
In this paper we propose an 
effective Unsupervised Domain Adaptation (UDA) strategy, based on a feature clustering method that captures the different semantic modes of the feature distribution and
groups features of the same class into tight and well-separated clusters. 
Furthermore, we introduce two novel learning objectives to enhance the discriminative clustering performance: 
an orthogonality loss forces spaced out individual representations to be orthogonal, while a sparsity loss reduces class-wise the number of active feature channels. 
The joint effect of these modules is to regularize the structure of the feature space.
%
%
Extensive evaluations in the synthetic-to-real scenario show that we achieve state-of-the-art performance.  
\end{abstract}


\section{Introduction}

\begin{figure}[htbp]
\includegraphics[width=0.95\linewidth]{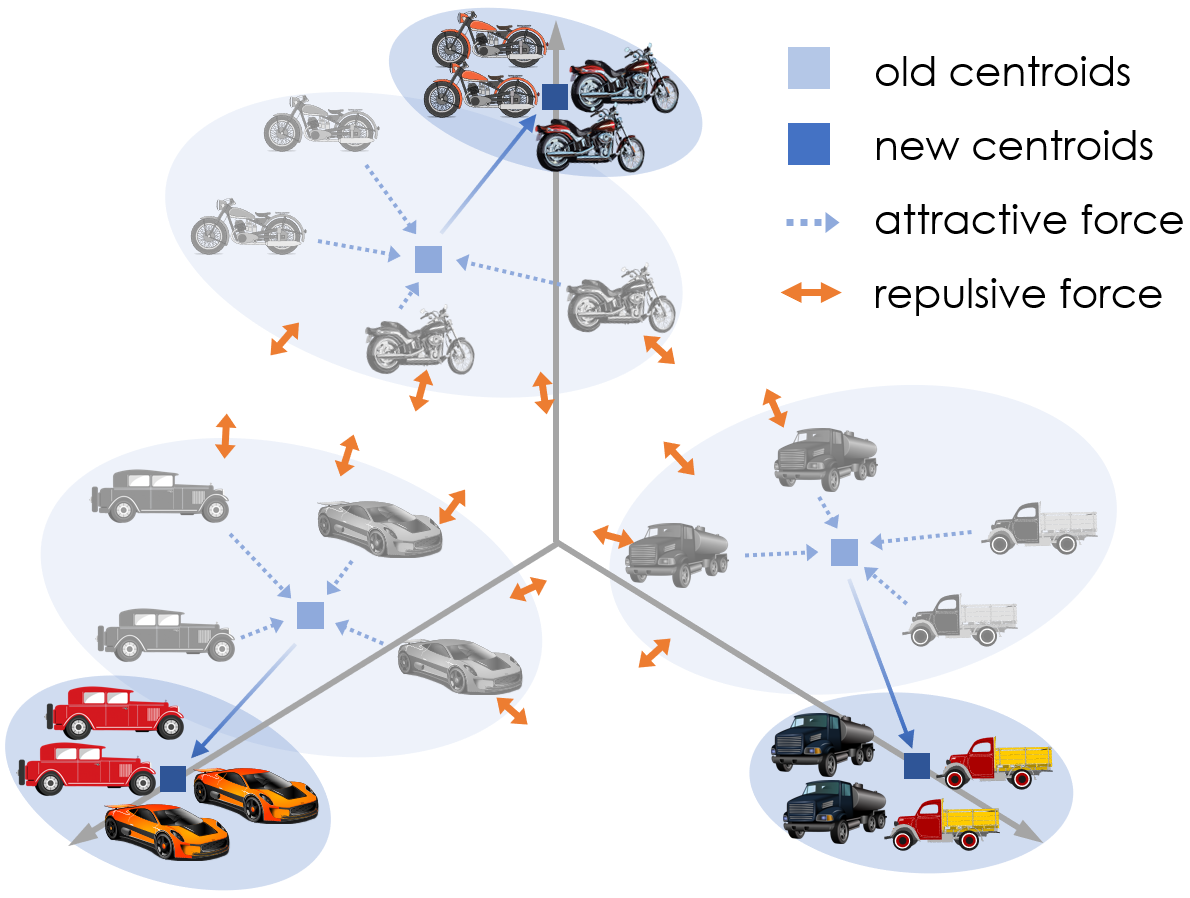}
\caption{The proposed domain adaptation scheme is driven by 3 main components, i.e., feature clustering, orthogonality and sparsity. These push features in the previous step (in light gray) to new locations (colored) where
features of the same class are clustered, while features of distinct classes are pushed away. To further improve performances, features of distinct classes are forced to be orthogonal and sparse.
}
\label{fig:graphical_abstract}
\end{figure}

Semantic segmentation  is one of the most challenging prediction tasks towards complete scene understanding and has achieved substantial improvements thanks to deep learning architectures.
State-of-the-art approaches typically rely on an auto-encoder structure, where an encoder extracts meaningful compact representations of the scene  and a decoder processes them to obtain a dense segmentation map.
Starting from the well-known FCN architecture \cite{long2015}, many models have been proposed, such as PSPNet \cite{zhao2017}, DRN \cite{yu2017drn} and DeepLab \cite{chen2018deeplab,chen2018encoder,chen2017rethinking}. 
The main drawback of such architectures is their high complexity: indeed, their success is strictly related to the availability of massive amounts of labeled data. For this reason, many datasets have been created (e.g., Cityscapes \cite{Cordts2016} or Mapillary \cite{neuhold2017mapillary} for urban scene understanding), but the pixel-wise annotation procedure is highly expensive and time consuming. 
To avoid this demanding process, UDA has come into play in order to exploit knowledge extracted from (related) data sources where labels are easily accessible to the problem at hand.

Three main levels to which adaptation may occur can be identified  \cite{toldo2020unsupervised}: namely, at the input, features or output stages. 
A popular solution has become to bridge the domain gap at an intermediate or output representation level by means of adversarial techniques \cite{hoffman2016,hoffman2018,michieli2020adversarial}. The major drawback of these kind of approaches is that they usually perform a semantically unaware alignment, as they neglect the underlying class-conditional data distribution. 
 Additionally, they typically require a long training time to converge and the process may be unstable.

Differently from these techniques, our  approach is simple and does not require complex adversarial learning schemes: 
it entails only a slight increase of computation time with respect to the sole supervised learning. 
In this work we focus on feature-level UDA.
The main idea is depicted in Fig.\ \ref{fig:graphical_abstract}: we devise a domain adaptation technique specifically targeted to guide the latent space organization and driven by 3 main components. The first is a  feature clustering method  developed to group together features of the same class, while pushing apart features belonging to different classes. This constraint, which works on both domains, is similar in spirit to the recent progresses in contrastive learning for classification problems \cite{chen2020simple}; however, it has been developed aiming at a simpler computation, as the number of features per image is significantly larger than in the classification task. The second is a novel orthogonality requirement for the feature space, aiming at reducing  the cross-talk between features belonging to different classes. Finally, a sparsity constraint is added to reduce the number of active channels for each feature vector: our aim is to enforce the capability of deep learning architectures to learn a compact representation of the scene. The combined effect of these modules allows to regularize the structure of the latent space in order to encompass the source and target domains in a shared representation.

%

Summarizing, our main contributions are:
(1) we extend feature clustering (similarly to contrastive learning)  to semantic segmentation; 
(2) we introduce orthogonality and sparsity objectives to force a regular structure of the embedding space;
(3) we achieve state-of-the-art results on feature-level adaptation on two widely used benchmarks.



\section{Related Work}
\label{sec:related}

\noindent
\revised{\textbf{Unsupervised Domain Adaptation.}}
%
The majority of the UDA techniques focus on a distribution matching to bridge the statistical gap affecting data representations from different domains. 
While early approaches directly quantify the domain shift in terms of first or second order statistics \cite{TzengHZSD14,long2015learning,LongZ0J16,SunFS16,SunS16}, 
extremely successful has been the introduction of adversarial learning 
as an adaptation technique \cite{ganin2015}:
a domain discriminator implicitly captures the domain discrepancy, 
which is then minimized in an adversarial fashion. 
Adversarial adaptation has been first introduced for image classification, both at pixel and feature levels, in order to produce a target-like artificial supervision \cite{LiuT16,bousmalis2017} 
or to learn a domain invariant latent space \cite{ganin2015,tzeng2017}. 
%
A variety of techniques have been proposed to tackle the UDA task, ranging from completely non-adversarial approaches \cite{Xie2018,Deng2019,Balaji2019,Xu2019,Cicek2019,Saito2019} to enhanced adversarial strategies \cite{cui2020gradually,wang2020progressive}.

\noindent
\textbf{UDA for Semantic Segmentation.} 
The aforementioned adaptation methods have been developed in the context of image classification. 
Nonetheless, semantic segmentation introduces further challenges, as dense classification demands for structured predictions that must capture high-level global contextual information, while simultaneously reaching pixel-level precision. 
Thus, domain adaptation in semantic segmentation calls for more sophisticated techniques to face the additional complexities. In turn, 
the adaptation is overly rewarding, being human effort in pixel-level labeling extremely expensive and time consuming. 
\\
Multiple methods have been recently proposed to tackle this scenario. 
Some works seek for statistical matching in the input (image) space to achieve cross-domain uniformity of visual appearance. This has been attained through generative adversarial techniques \cite{hoffman2018,MurezKKRK18,sankaranarayanan2018,chen2019crdoco,toldo2020,pizzati2020domain} and with style transfer solutions \cite{Dundar18,Wu19,Choi2019}. 
Alternatively, domain distribution alignment has been enforced over semantic representations, both on intermediate or output network levels. 
In this direction, adversarial learning has been widely adopted 
\revised{\cite{hoffman2016,Saito2018ADR,Park2018,Lee2019,tsai2018,Tsai2019,du2019ssfdan,paul2020domain}}, 
as well as entropy minimization \cite{Chen2019,vu2019advent}, self-training 
\revised{\cite{biasetton2019,michieli2020adversarial,spadotto2020unsupervised,zou2018,Zou2019,yang2020fda}}
or curriculum style approaches \cite{zhang2017,lian2019constructing}. 

\noindent
\textbf{Clustering in UDA.}
A few approaches have been proposed recently to address UDA in image classification by resorting to a discriminative clustering algorithm, whose  goal is to disclose target class-conditional modes in the feature space. 
\\
%
%
A group of works  \cite{kang2019contrastive,liang2019distant,wang2019unsupervised,tian2020domain} embed variations of the K-means 
algorithm in the overall adaptation framework, where clusters of geometrically close latent representations are identified, 
revealing the semantic modes from the unlabeled target domain. 
Being developed to address image classification, these clustering strategies may lose efficacy 
when dealing with semantic segmentation. 
Moreover, they deeply rely on a geometric measure of feature similarity to assign target pseudo-labels, which may not be feasible in a very high-dimensional space. 
For this reason, this type of clustering technique is often combined with a learnable projection to discover a more easily tractable lower dimensional latent space \cite{liang2019distant,wang2019unsupervised,tian2020domain}.
\\
To overcome the lack of target semantic supervision, other approaches \cite{Xie2018,Deng2019} resort to pseudo-labels directly from network predictions to discover target class-conditional structures in the feature space. Those structures are then exploited to perform a within-domain feature clusterization \cite{Deng2019} and cross-domain feature alignment by centroid matching \cite{Xie2018,Deng2019}. 
Starting from analogous premises, we extend a similar form of inter and intra class adaptation to the semantic segmentation scenario, by introducing additional modules that help to address the inherent increased complexity.
\\
%
Avoiding the need for target pseudo-labels, 
\revised{\cite{Saito2019, saito2020universal}} 
propose a self-supervised clustering technique to discover target modes without any form of supervision.
However, 
their approach is not easily scalable to semantic segmentation, as it requires to store feature embeddings of past samples, which is rather impractical when each data instance is associated with thousands of latent representations.
\\
%
Quite recently, Tang et al.\ \cite{tang2020unsupervised} argue that a direct class-wise alignment over source and target features could harm the discriminative structure of target data. Thus, they perform intrinsic feature alignment by a joint, yet distinct, model training with both source and target data. 
Nonetheless, in a more complex semantic segmentation scenario, an implicit adaptation could be not enough to bridge the domain gap that affects the effectiveness of target predictions.

\noindent
\textbf{Orthogonality and Sparsity.}
Deep neural networks are trained to learn a compact representation of the scene. However, no constraint is posed on the orthogonality among feature vectors belonging to different classes or on the sparsity of their activations \revised{\cite{shi2018entropy, wang2019clustering}}. 
The orthogonality between feature vectors has been recently proposed for UDA in \cite{pinheiro2018unsupervised,wu2019improving}. In these works, a penalty term is introduced to force the prototypes (i.e., the class centers) to be orthogonal. Differently from these approaches, we apply a feature-wise entropy minimization constraint, imposing that each feature is orthogonal with respect to all but one centroid.

To minimize the interference among features we drive them to be channel-wise sparse. To the best of our knowledge, there are no prior works employing channel-wise sparsification in deep learning models. However, some prior techniques exist for domain adaptation on linear models exploiting sparse codes on a shared dictionary between the domains \cite{shekhar2013generalized,zhang2015domain}. Additionally, in \cite{ranzato2007efficient} an expectation maximization approach is proposed to compute sparse code vectors which minimize the energy of the model. Although the approach is applied to linear encoders and decoders for simplicity, it could be extended to non-linear models.

\section{Method}

\begin{figure*}[t]
\centering
\includegraphics[width=0.95\linewidth]{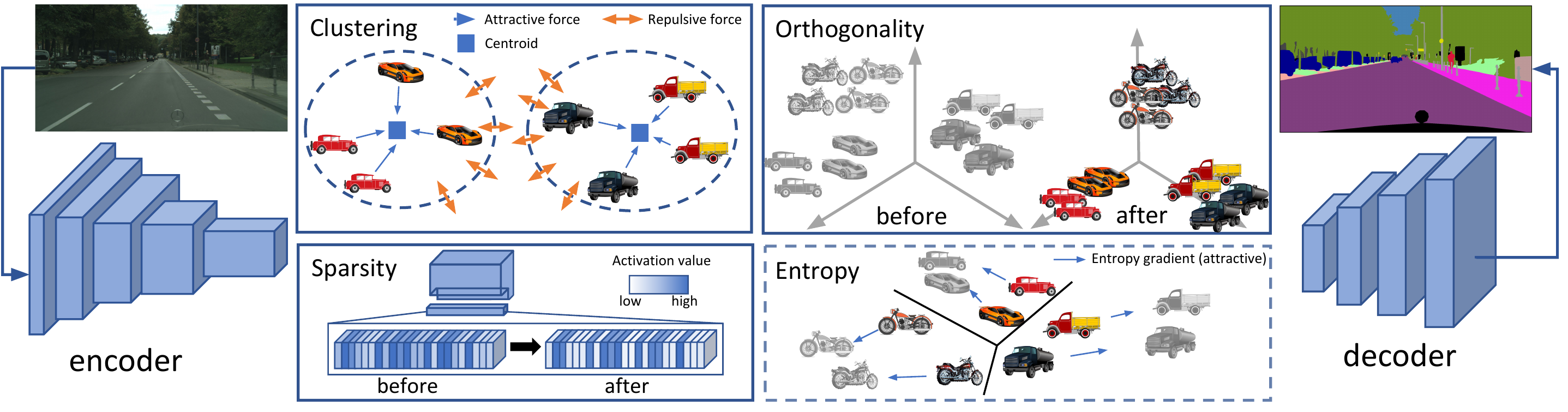}
\caption{Overview of the proposed approach. Features  after supervised training on the source domain are represented in light gray, while features of the current step are colored. 
A set of techniques is employed to better shape the latent feature space spanned by the encoder. Features are clustered and the clusters are forced to be disjoint. At the same time, features belonging to different classes are forced to be orthogonal with respect to each other. Additionally, features are forced to be sparse 
and an entropy minimization loss could also be added  to guide target samples far from the decision boundaries. 
}
\label{fig:architecture}
\end{figure*}

%
In this section, we provide an in depth description of the core modules of the proposed method. Our approach leverages a clustering objective applied over the individual feature representations, with novel orthogonality and sparsity constraints. 
Specifically, inter and intra class alignments are enforced by grouping together features of the same semantic class, while simultaneously pushing away those of different categories.
\revised{By enforcing the clustering objective on both source and target representations, we drive the model towards feature-level domain alignment.}
We further regularize the distribution of latent representations by the joint application of an orthogonality and a sparsity losses. 
The orthogonality module has a two-fold objective: first, it forces feature vectors of kindred semantic connotations to activate the same channels, while turning off the remaining ones; second, it constrains feature vectors of dissimilar semantic connotations to activate different channels, i.e., with no overlap, to reduce cross interference.
The sparsity objective further encourages a lower volume of active feature channels from latent representations, i.e., it concentrates the energy of the features on few dimensions. 
%

A graphical outline of the  approach with all its components is shown in Fig.\ \ref{fig:architecture}: the  training objective is given by the combination of the standard supervised loss with the proposed adaptation modules, i.e., it is computed as:
\begin{equation}
\mathcal{L}_{tot}' = \mathcal{L}_{ce} + \lambda_{cl} \cdot \mathcal{L}_{cl}
+ \lambda_{or} \cdot \mathcal{L}_{or} + \lambda_{sp} \cdot \mathcal{L}_{sp}
\label{eq:tot}
\end{equation}
where $\mathcal{L}_{ce}$ is the standard supervised cross entropy loss. The other components will be detailed in the following sections: the main clustering objective ($\mathcal{L}_{cl}$) is introduced in Section \ref{sec:clustering}. 
The orthogonality constraint ($\mathcal{L}_{or}$) is discussed in Section \ref{sec:orthogonality} and finally the sparsity constraint ($\mathcal{L}_{sp}$) is detailed in  Section \ref{sec:sparsity}. 
The $\lambda$ parameters balance the multiple losses and \revised{are} experimentally chosen using a validation set. \\
In addition, we further integrate the proposed adaptation method with an off-the-shelf entropy-minimization like objective ($\mathcal{L}_{em}$), to provide an extra regularizing action over the segmentation feature space and ultimately achieve an improved performance in some evaluation scenarios. In particular, we adopt the simple, yet effective, maximum squares objective of \cite{Chen2019}, in its \textit{image-wise class-balanced} version. 
Hence, we can define the ultimate training objective comprising the entropy module as:
\begin{equation}
\mathcal{L}_{tot} = \mathcal{L}_{tot}' + \lambda_{em} \cdot \mathcal{L}_{em}
\label{eq:tot2}
\end{equation}


\subsection{Discriminative Clustering}
\label{sec:clustering}

In the considered UDA setting,  we are provided with plenty of samples $\mathbf{X}_n^s \in \mathbb{R}^{H \times W \times 3}$ from a source dataset, 
in conjunction with their  semantic maps $\mathbf{Y}_n^s \in \mathbb{R}^{H \times W}$. Those semantic maps contain at each  spatial location a ground truth  index belonging to the set of possible classes $\mathcal{C}$, which denotes the semantic category of the associated pixel. 
Concurrently, we have at our disposal target training samples $\mathbf{X}_n^t \in \mathbb{R}^{H \times W \times 3}$ with no label maps (we allow only the availability of a small amount of target labels for validation and testing purposes).
Despite sharing similar high-level semantic content, the source and training samples are distributed differently, preventing a source-based model to achieve a satisfying prediction accuracy on target data without adaptation.  
We denote as $S=F \circ C$ the segmentation network composed of an encoder and a decoder modules, namely the feature extractor $F$ and the classifier $C$. Notice that the proposed method is agnostic to the employed deep learning model, except for the assumption of an auto-encoder structure
\revised{and of positive feature values as provided by ReLU activations that are typically placed at the encoder output} 
(as almost all the current state-of-the-art approaches for semantic segmentation).

To bridge the domain gap between the source and target datasets we operate at the feature level. 
The discrepancy of input statistics across domains is reflected into a shift of feature distribution in the latent space spanned by the feature extractor. This ultimately may cause the source-trained classifier to draw decision boundaries crossing high density regions of the target latent space \cite{vu2019advent}, since it is inherently unaware of the target semantic modes extracted from unlabeled target data. 
Thus, the classification performance over the target domain is strongly degraded when compared to the upper bound of the source prediction accuracy.  

We cope with this performance degradation by resorting to a clustering module, that serves as constraint towards a class-conditional feature alignment between domains. 
Given a batch of source ($\mathbf{X}_n^s$) and target ($\mathbf{X}_n^t$) training images (for ease of notation we pick a single image per domain), we first extract the feature tensors $\mathbf{F}_{n}^s = F(\mathbf{X}_n^s)$ and $\mathbf{F}_{n}^t = F(\mathbf{X}_n^t)$, along with the computed output segmentation  maps $\mathbf{S}_{n}^s = S(\mathbf{X}_n^s)$ and $\mathbf{S}_{n}^t = S(\mathbf{X}_n^t)$. 
The clustering loss is then computed as:
%
%
%
\begin{equation}
\mathcal{L}_{cl} \! = \! 
\frac{1}{|\mathbf{F}^{s,t}_n|} \sum_{\substack{\mathbf{f}_i \in \mathbf{F}^{s,t}_n \\ \hat y_i \in \mathbf{S}^{s,t}_n}} 
\!\!\! d(\mathbf{f}_i,\mathbf{c}_{\hat y_i})
- 
\frac{1}{|\mathcal{C}|(|\mathcal{C}| \! - \! 1)} 
\sum_{j \in \mathcal{C}} \sum_{\substack{k \in \mathcal{C} \\ k \neq j}} 
d(\mathbf{c}_{j},\! \mathbf{c}_{k})
\label{eq:L_cl}
\end{equation}
%
%
%
where $\mathbf{f}_i$ is an individual feature vector corresponding to a single spatial location from either source or target domain and  $\hat y_i$ is the corresponding predicted class (to compute  $\hat y_i$ the segmentation map  $\mathbf{S}_{n}^{s,t}$ is downsampled to match the feature tensor spatial dimensions). 
The function $d(\cdot)$ represents a generic distance measure, that we set to the $L1$ norm (we also tried the $L2$ norm but it yielded lower results). 
Finally, $\mathbf{c}_j$ denotes the centroid of semantic class $j \in \mathcal{C}$ computed according to the standard formula:
%
%
\begin{equation}
\mathbf{c}_j = 
\frac{ \sum_{\mathbf{f}_i}  \sum_{\hat y_i}  \delta_{j,\hat y_i}  \, \mathbf{f}_i }
{ \sum_{\hat y_i}   \delta_{j,\hat y_i} } 
, \quad j \in \mathcal{C}
\label{eq:centroid}
\end{equation}
where $\delta_{j,\hat y_i}$ is equal to $1$ if $\hat y_i = j$, and to $0$ otherwise.

The clustering objective is composed of two terms, the first measures how close features are from their respective centroids and the second how spaced out clusters corresponding to different semantic classes are.  
Hence, the effect provided by the loss minimization is twofold: firstly, feature vectors from the same class but  different domains are tightened around class feature centroids; secondly, features from separate classes are subject to a repulsive force applied to feature centroids, moving them apart. 

\subsection{Orthogonality}
\label{sec:orthogonality}
As opposed to previous works on clustering-based adaptation methods for image classification, in semantic segmentation 
additional complexity is brought by the dense structured classification. 
To this end, we first introduce an orthogonality constraint in the form of a training objective. 
More precisely, feature vectors from either domains, but of 
different semantic classes according to the network predictions, are forced to be orthogonal, meaning that their scalar product should be small. On the contrary, features sharing semantic classification should carry high similarity, i.e., large scalar product. 
Yet, feature tensors associated to training samples enclose thousands of feature vectors to cover the entire spatial extent of the scene and to reach pixel-level classification. 
Thus, since measuring pair-wise similarities requires a significant computational effort, we calculate the scalar product between each feature vector and every class centroid $\mathbf{c}_j$  (centroids are computed using Eq.~\ref{eq:centroid}). 
\revised{Inspired by \cite{Saito2019,saito2020universal}, we devise the orthogonality objective as 
an entropy minimization loss that forces each feature to be orthogonal with respect to all the centroids but one: 
}
%
\begin{equation}
\mathcal{L}_{or} =  - \! \! \!
\sum_{\mathbf{f}_i \in F(\mathbf{X}^{s,t}_n)} \sum_{j \in \mathcal{C}} 
p_{j} (\mathbf{f}_i) \log p_{j} (\mathbf{f}_i)
\label{eq:L_orth}
\end{equation}
where $\{p_{j}(\mathbf{f}_i)\}$ denotes a probability distribution derived as: 
%
%
\begin{equation}
p_{j} (\mathbf{f}_i)  = \frac{e^{\langle \mathbf{f}_i , \mathbf{c}_j \rangle}}
{\sum_{k \in \mathcal{C}} e^{\langle \mathbf{f}_i , \mathbf{c}_k \rangle}}
, \quad j \in \mathcal{C}
\label{eq:p}
\end{equation}
The loss minimization forces a peaked distribution of the probabilities $\{p_{j}(\mathbf{f}_i)\}$,  promoting the orthogonality property as described above, since each feature vector is compelled to carry a high similarity score with a single class centroid. 
The overall effect of the orthogonality objective is to promote a regularized feature distribution, which should ultimately boost the clustering efficacy in performing domain feature alignment.

\subsection{Sparsity}
\label{sec:sparsity}
To strengthen the regularizing effect brought by the orthogonality constraint, we introduce a further training objective to better shape class-wise feature structures inside the latent space. 
In particular, we propose a sparsity loss, with the intent of decreasing the number of active feature channels of latent vectors. 
The objective is defined as follows: 
%
%
%
\begin{equation}
\mathcal{L}_{sp} = -
\sum_{i \in \mathcal{C}} 
|| \tilde{\mathbf{c}}_{i} - \bm{\rho}  ||_2^2
\label{eq:L_spa}
\end{equation}
where $\tilde{\mathbf{c}}_{i}$ stands for the normalized centroid $\mathbf{c}_{i}$ in $[0,1]^D$ and $D$ denotes the number of feature maps in the encoder output. 
We also empirically set $\bm{\rho} = [0.5]^D$. 
It can be noted that the sparsifying action is delivered on class centroids, thus applying an indirect, yet homogeneous, influence over all feature vectors from the same semantic category. 
The result is a semantically-consistent suppression of weak activations, while rather active ones are jointly raised.

While the orthogonality objective aims at promoting different sets of activations on feature vectors from separate semantic classes, the sparsity loss seeks to narrow those sets to a limited amount of units.   
Again, the  goal is to ease the clustering loss task in creating tight and well distanced aggregations of features of similar semantic connotation from either source and target domains, by providing an improved regularity to the class-conditional semantic structures inside the feature space.

\section{Experimental Setup}

\textbf{Datasets.} 
Supervised training is performed on the synthetic datasets GTA5 \cite{Richter2016} and  SYNTHIA \cite{ros2016}. We employ Cityscapes \cite{Cordts2016} as target domain. 
The GTA5 \cite{Richter2016} contains $24966$ synthetic images taken from a car perspective in US-style virtual cities, with a high level of quality and realism. 
On the other hand, the SYNTHIA \cite{ros2016} offers $9400$ images from virtual European-style towns with large scene variability under various light and weather conditions, but little visual quality. 
For evaluation on real datasets, respectively $19$ and $16$ compatible classes are considered when adapting from GTA5 and SYNTHIA.
The Cityscapes \cite{Cordts2016} is instead provided with the limited amount of $2975$ images acquired in $50$ European cities.  
The original training set (without labels) is used for unsupervised adaptation, while the $500$ images in the original validation set are used as a test set.


\textbf{Model Architecture.} The modules introduced in this work are agnostic to the underlying network architecture and can be extended to other scenarios. For fair comparison with previous works \cite{tsai2018, Chen2019,vu2019advent} we employ the DeepLab-V2, a fully convolutional segmentation network with ResNet-101 \cite{he2016deep} or VGG-16 \cite{simonyan2015very} as backbones. 
Further details on the segmentation network architecture can be found in \cite{tsai2018,vu2019advent}, as we follow the same implementation adopted in those works. 
We initialize the two encoder networks with ImageNet \cite{deng2009imagenet} pretrained weights. 
In addition, prior to the actual adaptation phase, we supervisedly train the segmentation network on source data.  

\textbf{Training Details.} The model is trained with the starting learning rate set to $2.5 \times 10^{-4}$ and decreased with a polynomial decay rule of power $0.9$. 
We employ weight decay regularization of $5 \times 10^{-4}$. 
Following \cite{Chen2019}, we also randomly apply  mirroring and gaussian blurring for data augmentation during the training stage.
To accommodate for GPU memory limitations, we resize images from the GTA5 dataset up to a resolution of $1280 \times 720\ \mathrm{px}$, as done by \cite{tsai2018}. 
SYNTHIA images are instead kept to the original size of $1280 \times 780\ \mathrm{px}$. 
As for the target Cityscapes dataset, training unlabeled images are resized to $1024 \times 512\ \mathrm{px}$, whereas the results of the testing stage are reported at the original image resolution ($2048 \times 1024\ \mathrm{px}$). 
We use a batch size of 1 and the hyper-parameters are tuned by resorting to a small subset of labeled target data that we set aside from the original target training set and reserve only for parameter selection. 
As evaluation metric, we employ the mean Intersection over Union (mIoU).  
The entire model is developed using PyTorch and trained with a single GPU. 
\revised{The code is available at \url{https://lttm.dei.unipd.it/paper_data/UDAclustering/}.} 


\section{Results}

\newcommand{\mcrot}[3]{\multicolumn{#1}{#2}{\rlap{\rotatebox{50}{#3}~}}}

\renewcommand{\arraystretch}{0.67}
\begin{table*}[htbp]
    \centering
		\footnotesize
		\setlength{\tabcolsep}{1.6pt} 
    
		\makebox[\textwidth]{
		
\begin{tabular}{c @{\hspace{4mm}} c @{\hspace{3mm}} c @{\hspace{1.7mm}}|@{\hspace{1.7mm}} *{19}c @{\hspace{1.7mm}}| c | c}

\toprule

    & \multicolumn{1}{c}{B \phantom{B}} & \multicolumn{1}{c}{Method} 
    & \mcrot{1}{l}{Road} & \mcrot{1}{l}{Sidewalk} & \mcrot{1}{l}{Building}  & \mcrot{1}{l}{Wall\textsuperscript{*}}       & \mcrot{1}{l}{Fence\textsuperscript{*}} 
    & \mcrot{1}{l}{Pole\textsuperscript{*}} & \mcrot{1}{l}{T. Light} & \mcrot{1}{l}{T. Sign}   & \mcrot{1}{l}{Vegetation} & \mcrot{1}{l}{Terrain}
	& \mcrot{1}{l}{Sky}  & \mcrot{1}{l}{Person}   & \mcrot{1}{l}{Rider}     & \mcrot{1}{l}{Car}        & \mcrot{1}{l}{Truck}
	& \mcrot{1}{l}{Bus}  & \mcrot{1}{l}{Train}    & \mcrot{1}{l}{Motorbike} & \mcrot{1}{l}{Bicycle}    & \multicolumn{1}{c}{{\makecell[b]{mIoU \\ (all) }}} & \multicolumn{1}{c}{\makecell[b]{mIoU\textsuperscript{*} \\ (13-cl)}}\\
	
	\addlinespace[0.5mm]
    \toprule
        
        \multirow{14}{*}{\rotatebox{90}{GTA5 $\rightarrow$ Cityscapes \hspace{0.9cm}}} & \multirow{10}{*}{\rotatebox{90}{VGG16}}
		& Source Only                   & 26.5 & 13.3 & 45.1 &  6.0 & 15.2 & 16.5 & 21.3 &  8.5 & 78.0 &  8.3 & 59.7 & 45.0 & 10.5 & 69.1 & 22.8 & 17.9 & 0.0 & 16.4 & 2.7 & 25.4 & - \\
		\cmidrule{3-24}
		& & FCNs ITW \cite{hoffman2016}       & 70.4 & \textbf{32.4} & 62.1 & 14.9 &  5.4 & 10.9 & 14.2 &  2.7 & 79.2 & 21.3 & 64.6 & 44.1 &  4.2 & 70.4 &  8.0 &  7.3 & 0.0 &  3.5 & 0.0 & 27.1 & -  \\
		& & CyCADA (feat) \cite{hoffman2018}  & 85.6 & 30.7 & 74.7 & 14.4 & 13.0 & 17.6 & 13.7 &  5.8 & 74.6 & 15.8 & 69.9 & 38.2 &  3.5 & 72.3 & 16.0 &  5.0 & 0.1 &  3.6 & 0.0 & 29.2 & -  \\
		& & CBST \cite{zou2018}               & 66.7 & 26.8 & 73.7 & 14.8 &  9.5 & \textbf{28.3} & 25.9 & 10.1 & 75.5 & 15.7 & 51.6 & \textbf{47.2} &  6.2 & 71.9 &  3.7 &  2.2 & \textbf{5.4} & \textbf{18.9} & \textbf{32.4} & 30.9 & -  \\
		& & MinEnt \cite{vu2019advent}        & 85.1 & 18.9 & 76.3 & \textbf{32.4} & \textbf{19.7} & 19.9 & 21.0 &  8.9 & 76.3 & \textbf{26.2} & 63.1 & 42.8 &  5.9 & \textbf{80.8} & \textbf{20.2} &  9.8 & 0.0 & 14.8 & 0.6 & 32.8 & -  \\
		& & MaxSquare IW \textsuperscript{(r)} \cite{Chen2019} & 81.4 & 20.0 & 75.4 & 19.4 & 19.1 & 16.1 & 24.4 & 7.9 & 78.8 & 22.9 & 65.9 & 45.0 & 12.3 & 74.6 & 16.1 & 10.3 & 0.2 & 11.3 & 1.0 & 31.7 & - \\
		& & Ours ($\mathcal{L}_{tot}'$)    & 83.6 & 16.6 & 79.0 & 19.8 & 18.7 & 21.5 & 27.3 & \textbf{15.9} & 80.2 & 14.3 & \textbf{72.6} & 47.0 & 17.5 & 76.8 & 16.6 & \textbf{13.9} & 0.1 & 16.0 & 3.4 & 33.7 & -  \\
		& & Ours ($\mathcal{L}_{tot}$)    & \textbf{86.0} & 13.5 & \textbf{79.4} & 20.4 & 18.5 & 21.5 & \textbf{27.6} & 15.2 & \textbf{80.8} & 21.9 & \textbf{72.6} & 46.3 & \textbf{18.1} & 80.0 & 16.9 & 13.1 & 1.0 & 14.6 & 2.0 & \textbf{34.2} & -  \\
		
		\cmidrule{2-24}
		
		& \multirow{9}{*}{\rotatebox{90}{ResNet101}}
		& Source Only                   & 81.8 & 16.3 & 74.4 & 18.6 & 12.7 & 23.5 & 29.3 & 18.1 & 73.5 & 21.4 & 77.6 & 55.6 & 25.6 & 74.1 & 28.6 & 10.2 & 3.0 & 25.8 & 32.7 & 37.0 & -  \\
		\cmidrule{3-24}		
		& & AdaptSegNet (feat) \cite{tsai2018} & 83.7 & 27.6 & 75.5 & 20.3 & 19.9 & 27.4 & 28.3 & 27.4 & 79.0 & 28.4 & 70.1 & 55.1 & 20.2 & 72.9 & 22.5 & 35.7 & \textbf{8.3} & 20.6 & 23.0 & 39.3 & -  \\
		& & MinEnt \cite{vu2019advent}         & 84.4 & 18.7 & 80.6 & 23.8 & 23.2 & 28.4 & 36.9 & 23.4 & 83.2 & 25.2 & \textbf{79.4} & 59.0 & \textbf{29.9} & 78.5 & 33.7 & 29.6 & 1.7 & 29.9 & 33.6 & 42.3 & -  \\
		& & SAPNet \cite{li2020spatial}        & 88.4 & 38.7 & 79.5 & \textbf{29.4} & \textbf{24.7} & 27.3 & 32.6 & 20.4 & 82.2 & 32.9 & 73.3 & 55.5 & 26.9 & 82.4 & 31.8 & 41.8 & 2.4 & 26.5 & 24.1 & 43.2 & -  \\
		& & MaxSquare IW \cite{Chen2019}       & 89.3 & \textbf{40.5} & 81.2 & 29.0 & 20.4 & 25.6 & 34.4 & 19.0 & 83.6 & 34.4 & 76.5 & 59.2 & 27.4 & 83.8 & \textbf{38.4} & 43.6 & 7.1 & \textbf{32.2} & 32.5 & 45.2 & -  \\
        & & Ours ($\mathcal{L}_{tot}'$)    & 88.7 & 32.2 & 81.8 & 24.1 & 22.1 & \textbf{30.8} & \textbf{37.6} & \textbf{32.8} & 83.4 & 36.3 & 76.0 & 60.0 & 27.0 & 81.0 & 34.2 & 43.0 & 8.0 & 23.4 & 38.1 & 45.3 & -  \\
		& & Ours ($\mathcal{L}_{tot}$)    & \textbf{89.4} & 30.7 & \textbf{82.1} & 23.0 & 22.0 & 29.2 & \textbf{37.6} & 31.7 & \textbf{83.9} & \textbf{37.9} & 78.3 & \textbf{60.7} & 27.4 & \textbf{84.6} & 37.6 & \textbf{44.7} & 7.3 & 26.0 & \textbf{38.9} & \textbf{45.9} & -  \\


    \midrule

        \multirow{14}{*}{\rotatebox{90}{SYNTHIA $\rightarrow$ Cityscapes \hspace{0.7cm}}} & \multirow{10}{*}{\rotatebox{90}{VGG16}}
		& Source Only  & 7.8 & 13.7 & 66.6 & 2.2 & 0.0 & 23.9 & 4.8 & 13.3 & 71.2  & - & 76.5 & 49.2 & 12.1 & 67.1  & - & 24.5  & - & 9.8 & 9.2 & 28.3 & 32.8\\
		\cmidrule{3-24}
		& & FCNs ITW \cite{hoffman2016}      & 11.5 & 19.6 & 30.8 &  4.4 &  0.0 & 20.3 & 0.1  & 11.7 & 42.3  & - & 68.7 & 51.2 &  3.8 & 54.0  & - &  3.2  & - & 0.2 &  0.6 & 20.2 & 22.9 \\
		& & Cross-City \cite{chen2017nomore} & 62.7 & 25.6 & \textbf{78.3} &    - &   -  &   -  & 1.2  &  5.4 & 81.3  & - & 81.0 & 37.4 &  6.4 & 63.5  & - & 16.1  & - & 1.2 &  4.6 &    - & 35.7 \\
		& & CBST \cite{zou2018}              & 69.6 & 28.7 & 69.5 & \textbf{12.1} &  \textbf{0.1} & \textbf{25.4} & 11.9 & 13.6 & \textbf{82.0}  & - & \textbf{81.9} & 49.1 & 14.5 & 66.0  & - &  6.6  & - & 3.7 & \textbf{32.4} & 35.4 & 36.1 \\
		& & MinEnt \cite{vu2019advent}       & 37.8 & 18.2 & 65.8 &  2.0 &  0.0 & 15.5 &  0.0 &  0.0 & 76.0  & - & 73.9 & 45.7 & 11.3 & 66.6  & - & 13.3  & - & 1.5 & 13.1 & 27.5 & 32.5 \\
		& & MaxSquare IW \textsuperscript{(r)}  \cite{Chen2019} & 9.1 & 12.7 & 72.5 & 1.0 & 0.0 & 22.3 & 7.0 & 8.4 & 80.0 & - & 77.9 & 49.4 & 10.0 & 71.8 & - & 23.8 & - & 6.0 & 13.5 & 29.1 & 34.0 \\
		& & Ours ($\mathcal{L}_{tot}'$)  & \textbf{78.5} & 29.9 & 77.7 & 1.2 & \textbf{0.1} & 24.1 & 11.9 & \textbf{15.0} & 78.7  & - & 78.5 & 51.0 & 15.4 & 73.7  & - & \textbf{24.7}  & - & \textbf{10.1} & 23.5 & \textbf{37.1} & \textbf{43.7} \\
		& & Ours ($\mathcal{L}_{tot}$)  & 78.3 & \textbf{30.1} & 78.0 & 1.7 & \textbf{0.1} & 24.1 & \textbf{12.0} & 14.6 & 79.7  & - & 79.1 & \textbf{51.4} & \textbf{15.5} & \textbf{74.4}  & - & 23.7  & - & 9.1 & 22.7 & \textbf{37.1} & \textbf{43.7} \\
		
		\cmidrule{2-24}
		
		& \multirow{9}{*}{\rotatebox{90}{ResNet101}}
		& Source Only                   & 39.5 & 18.1 & 75.5 & 10.5 & 0.1 & 26.3 & 9.0 & 11.7 & 78.6  & - & 81.6 & 57.7 & 21.0 & 59.9  & - & 30.1  & - & 15.7 & 28.2 & 35.2 & 40.5\\
		\cmidrule{3-24}
		& & AdaptSegNet (feat) \cite{tsai2018}  & 62.4 & 21.9 & 76.3 &   - &   - &    - & 11.7 & 11.4 & 75.3  & - & 80.9 & 53.7 & 18.5 & 59.7  & - & 13.7  & - & \textbf{20.6} & 24.0 & - & 40.8 \\
		& & MinEnt \cite{vu2019advent}          & 73.5 & 29.2 & 77.1 & 7.7 & 0.2 & 27.0 &  7.1 & 11.4 & 76.7  & - & 82.1 & 57.2 & \textbf{21.3} & 69.4  & - & 29.2  & - & 12.9 & 27.9 & 38.1 & 44.2 \\
		& & SAPNet \cite{li2020spatial}         & 81.7 & 33.5 & 75.9 &   - &   - &    - &  7.0 &  6.3 & 74.8  & - & 78.9 & 52.1 & \textbf{21.3} & 75.7  & - & \textbf{30.6}  & - & 10.8 & 28.0 &    - & 44.3 \\
		& & MaxSquare IW \cite{Chen2019}        & 78.5 & 34.7 & 76.3 & 6.5 & 0.1 & \textbf{30.4} & 12.4 & 12.2 & \textbf{82.2}  & - & \textbf{84.3} & \textbf{59.9} & 17.9 & 80.6  & - & 24.1  & - & 15.2 & 31.2 & 40.4 & 46.9  \\
		& & Ours ($\mathcal{L}_{tot}'$)    & 64.4 & 25.5 & 77.3 & \textbf{14.3} & \textbf{0.9} & 29.6 & \textbf{21.2} & \textbf{24.2} & 76.6  & - & 79.7 & 53.7 & 15.5 & 79.7  & - & 11.0  & - & 11.0 & \textbf{35.2} & 38.7 & 44.2 \\	
		& & Ours ($\mathcal{L}_{tot}$)    & \textbf{88.3} & \textbf{42.2} & \textbf{79.1} & 7.1 & 0.2 & 24.4 & 16.8 & 16.5 & 80.0  & - & \textbf{84.3} & 56.2 & 15.0 & \textbf{83.5}  & - & 27.2  & - & 6.3 & 30.7 & \textbf{41.1} & \textbf{48.2} \\	
	
    \bottomrule
        
\end{tabular}
}
\caption{Numerical evaluation of the GTA5 and SYNTHIA to Cityscapes adaptation scenarios in terms of per-class and mean IoU. Evaluations are performed on the validation set of the Cityscapes dataset. In all the experiments the DeepLab-V2 segmentation network is employed, with VGG-16 (top) or ResNet-101 (bottom) backbones. The mIoU\textsuperscript{*} results in the last column refer to the 13-classes configuration, i.e., classes marked with $^*$ are ignored. MaxSquares IW \textsuperscript{(r)} denotes our re-implementation, 
as original results are provided only for the ResNet-101 backbone.}
\label{tab:all}
\end{table*}
\newcommand{\imgsize}{28.5mm}
\begin{figure*}[htbp]
\centering
\begin{subfigure}[htbp]{\textwidth}
\hspace{0.08cm}
\centering
\resizebox{0.984\textwidth}{!}{%
\renewcommand{\arraystretch}{0.97}
\scriptsize
\begin{tabular}{@{}cccccccccc}

\cellcolor[HTML]{804080}{\color[HTML]{FFFFFF} \textbf{road}} & \cellcolor[HTML]{F423E8}\textbf{sidewalk} & \cellcolor[HTML]{464646}{\color[HTML]{FFFFFF} \textbf{building}} & \cellcolor[HTML]{66669C}{\color[HTML]{FFFFFF} \textbf{wall}} & \cellcolor[HTML]{BE9999}\textbf{fence} & \cellcolor[HTML]{999999}\textbf{pole} & \cellcolor[HTML]{FAAA1E}\textbf{traffic light} & \cellcolor[HTML]{DCDC00}\textbf{traffic sign} &\cellcolor[HTML]{6B8E23} \textbf{vegetation} & \cellcolor[HTML]{98FB98}\textbf{terrain} \\ 
\cellcolor[HTML]{4682B4}\textbf{sky} & \cellcolor[HTML]{DC143C}{\color[HTML]{FFFFFF} \textbf{person}} & \cellcolor[HTML]{FF0000}{\color[HTML]{FFFFFF} \textbf{rider}} & \cellcolor[HTML]{00008E}{\color[HTML]{FFFFFF} \textbf{car}} & \cellcolor[HTML]{000046}{\color[HTML]{FFFFFF} \textbf{truck}} & \cellcolor[HTML]{003C64}{\color[HTML]{FFFFFF} \textbf{bus}} & \cellcolor[HTML]{005064}{\color[HTML]{FFFFFF} \textbf{train}} & \cellcolor[HTML]{0000E6}{\color[HTML]{FFFFFF} \textbf{motorcycle}} & \cellcolor[HTML]{770B20}{\color[HTML]{FFFFFF} \textbf{bicycle}} & \cellcolor[HTML]{000000}{\color[HTML]{FFFFFF} \textbf{unlabeled}} \\ 
\end{tabular}%
}

\vspace{0.05cm}
\end{subfigure}
\setlength{\tabcolsep}{0.7pt} 
\renewcommand{\arraystretch}{0.6}
\centering
\begin{subfigure}[htbp]{\textwidth}
\centering
\begin{tabular}{cccccc}

   \includegraphics[width=\imgsize]{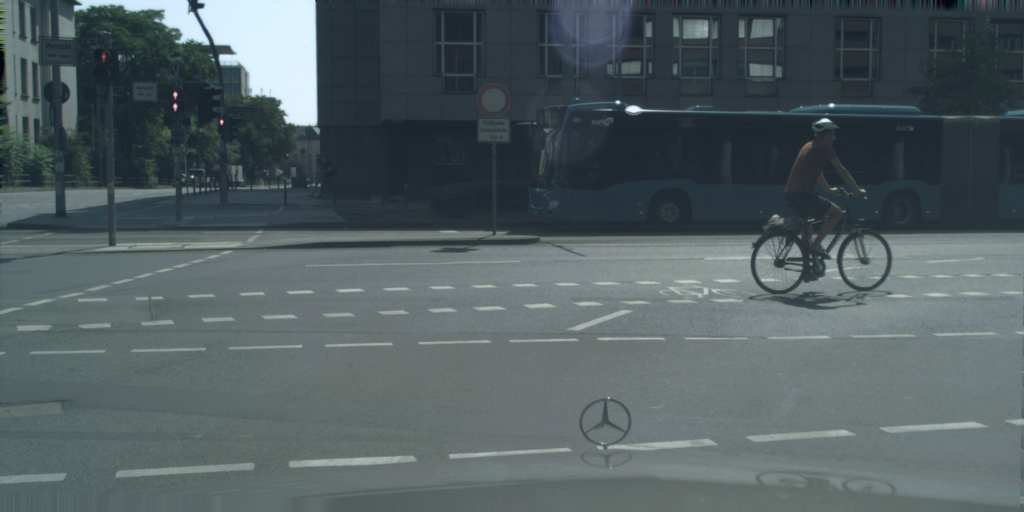} &
   \includegraphics[width=\imgsize]{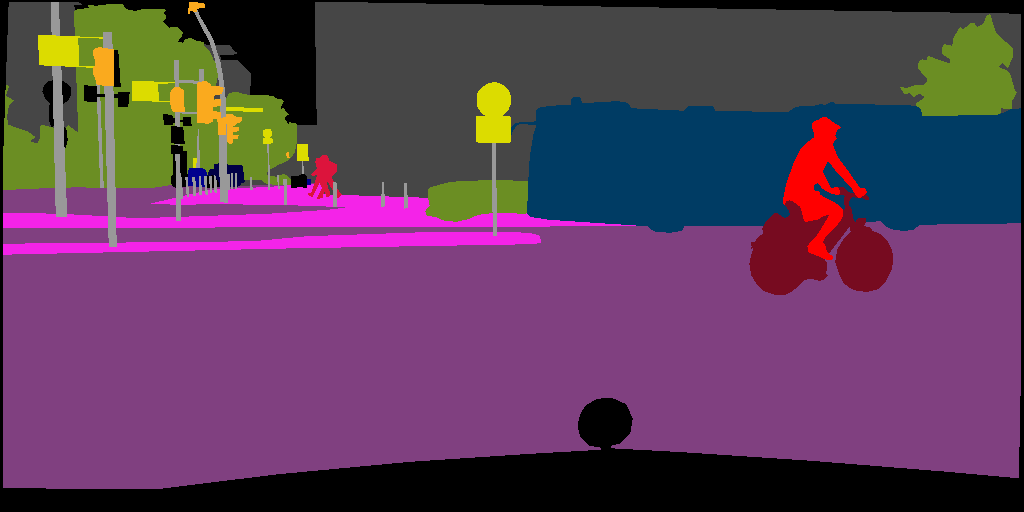} & 
   \includegraphics[width=\imgsize]{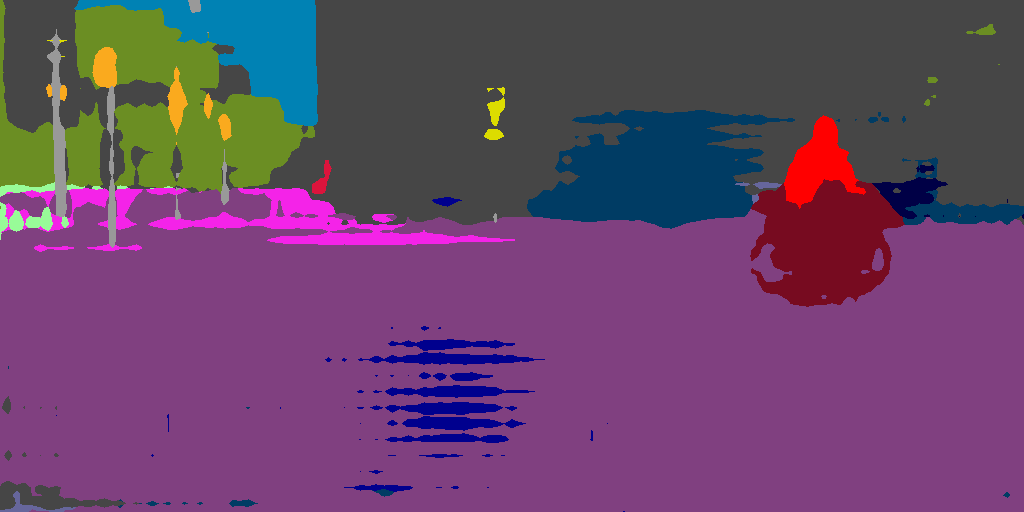} & 
   \includegraphics[width=\imgsize]{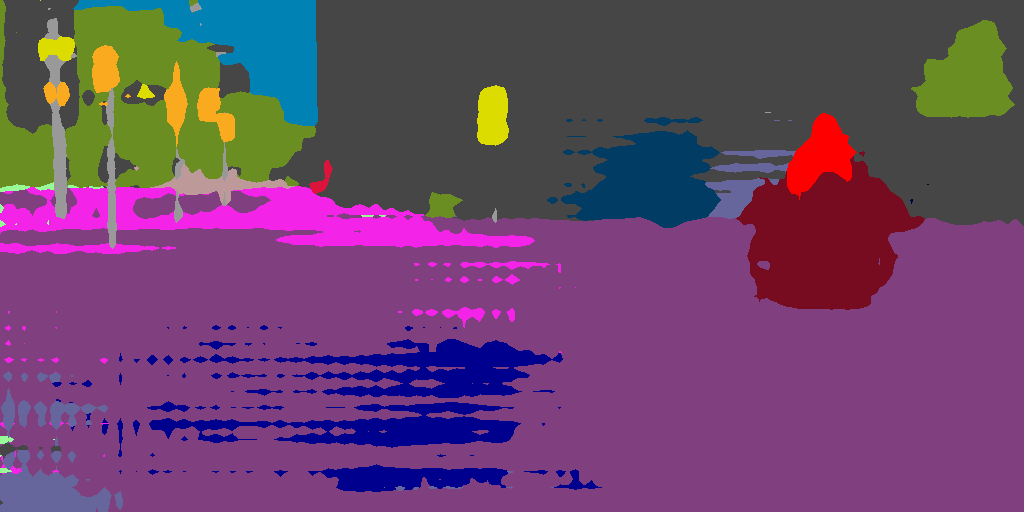} & 
   \includegraphics[width=\imgsize]{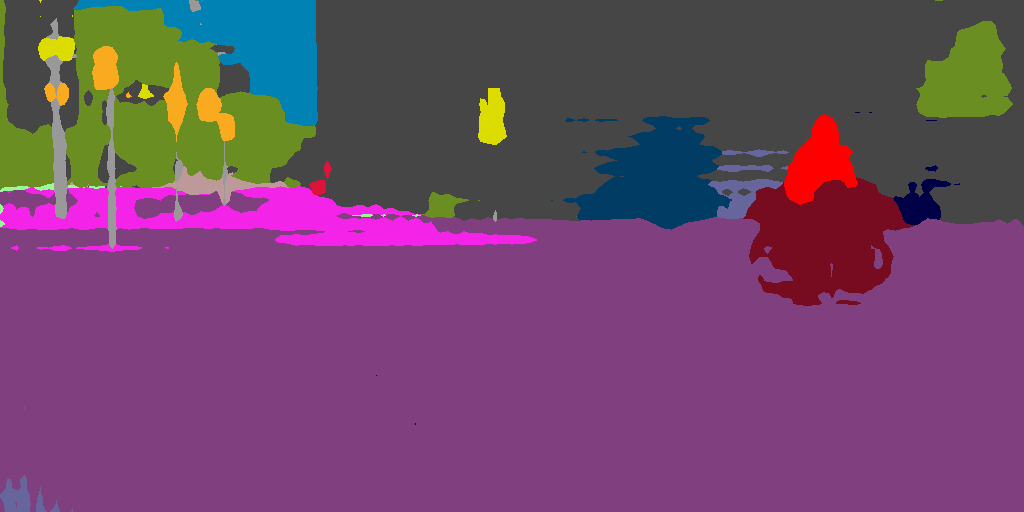} &
   \includegraphics[width=\imgsize]{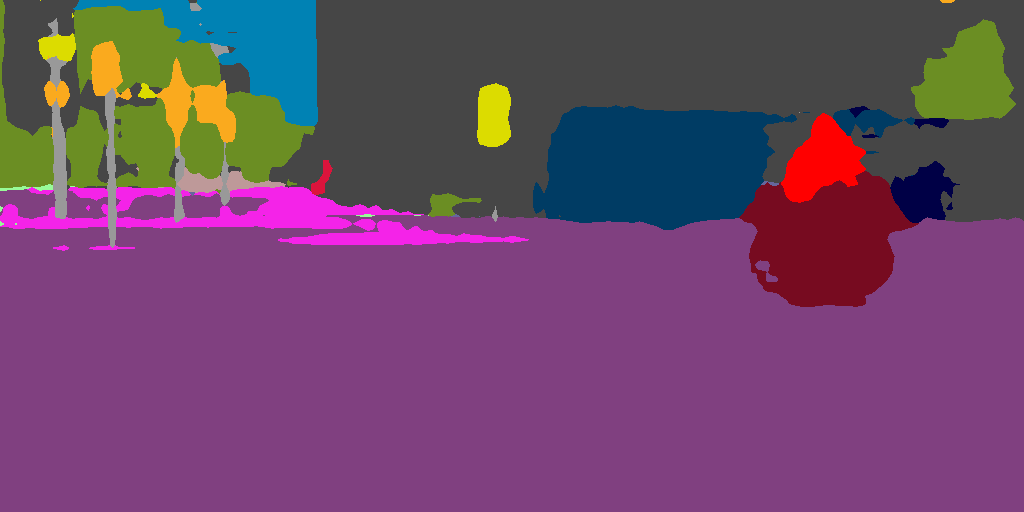} \\ 
   
   \includegraphics[width=\imgsize]{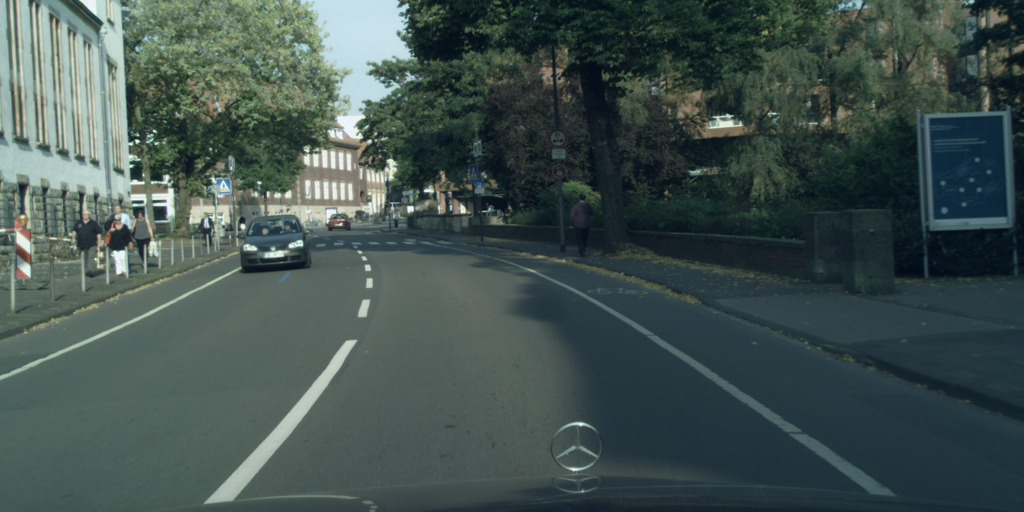} &
   \includegraphics[width=\imgsize]{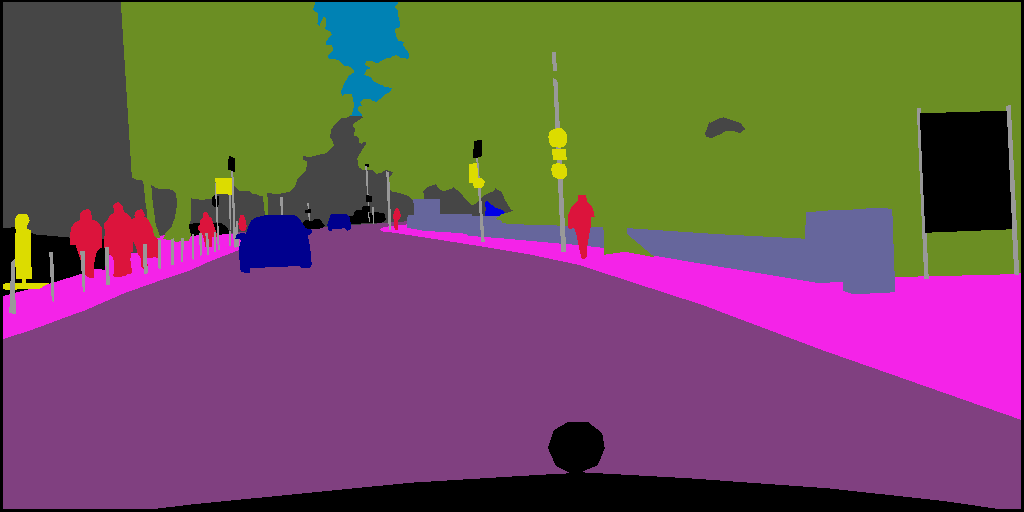} & 
   \includegraphics[width=\imgsize]{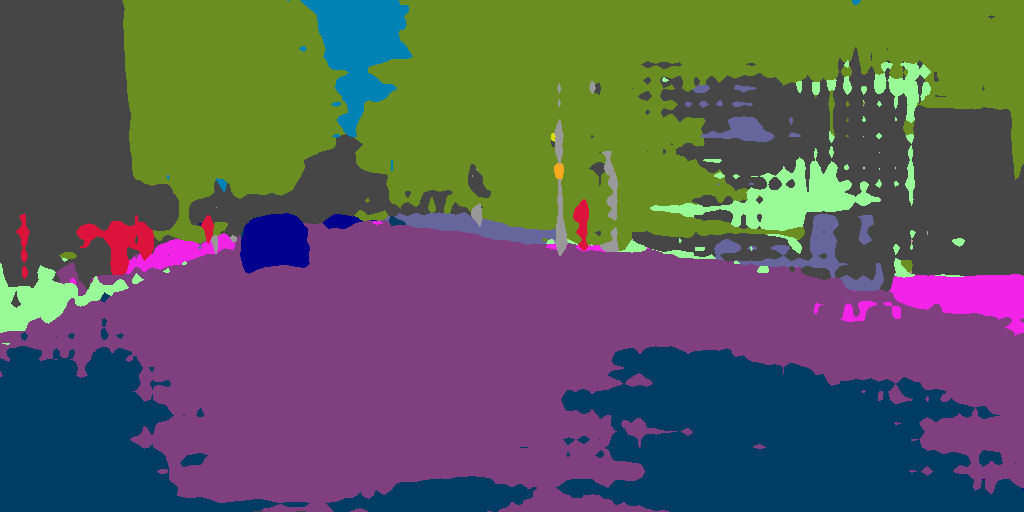} & 
   \includegraphics[width=\imgsize]{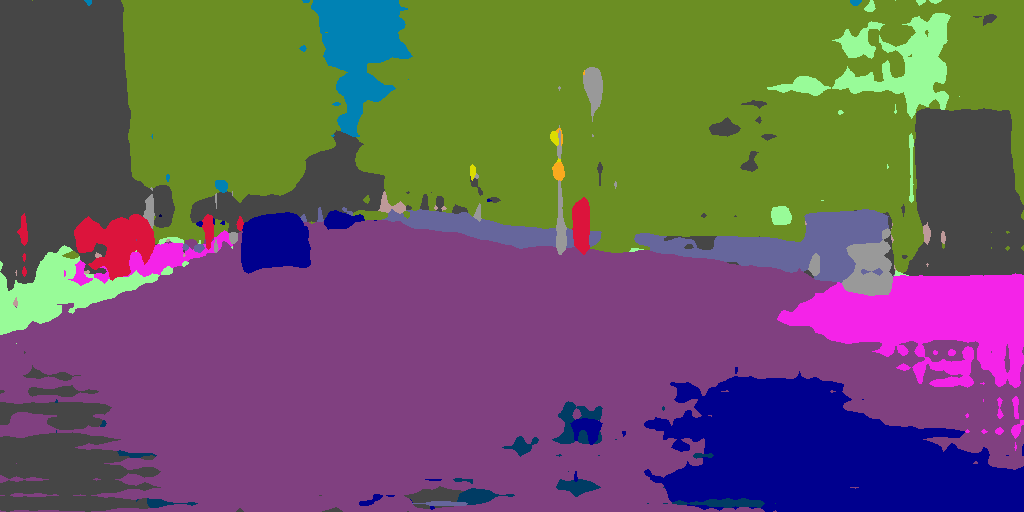} & 
   \includegraphics[width=\imgsize]{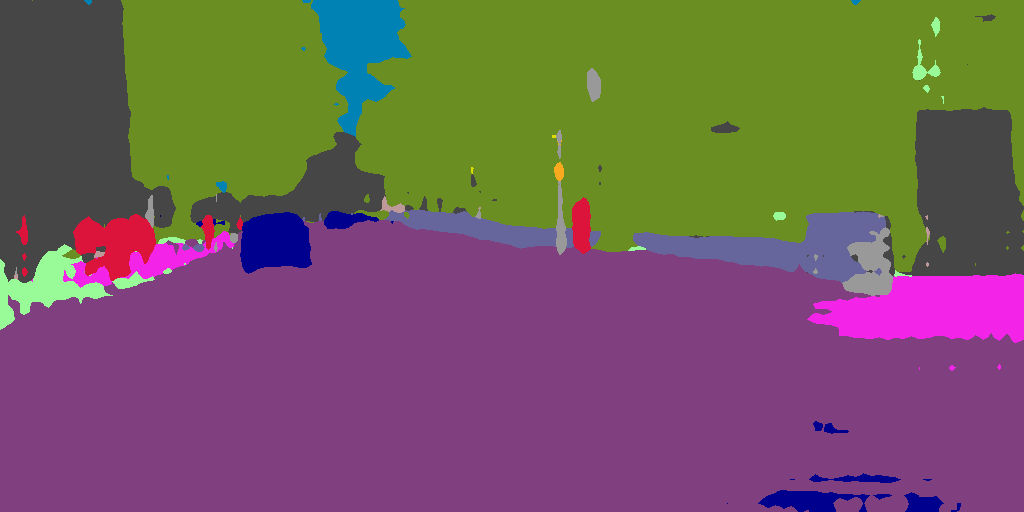} &
   \includegraphics[width=\imgsize]{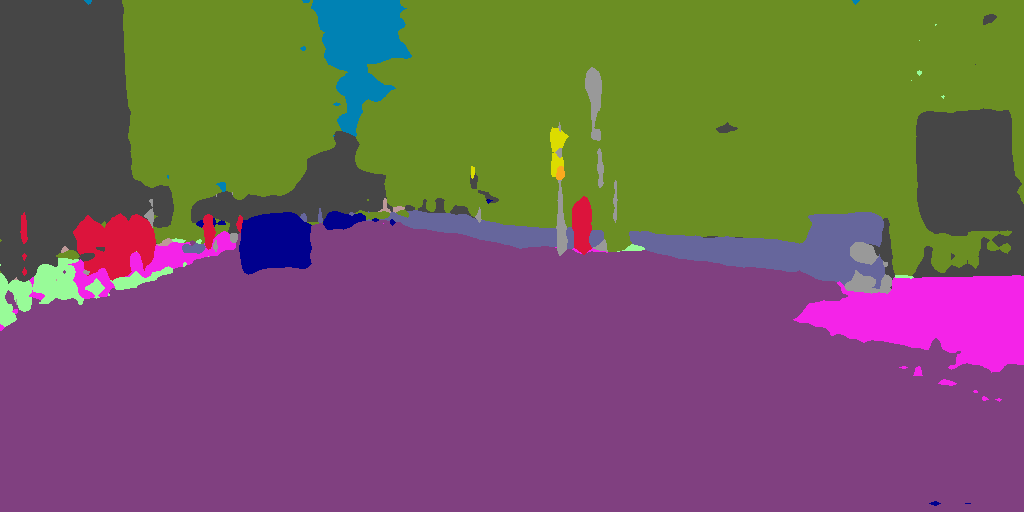} \\    

   \includegraphics[width=\imgsize]{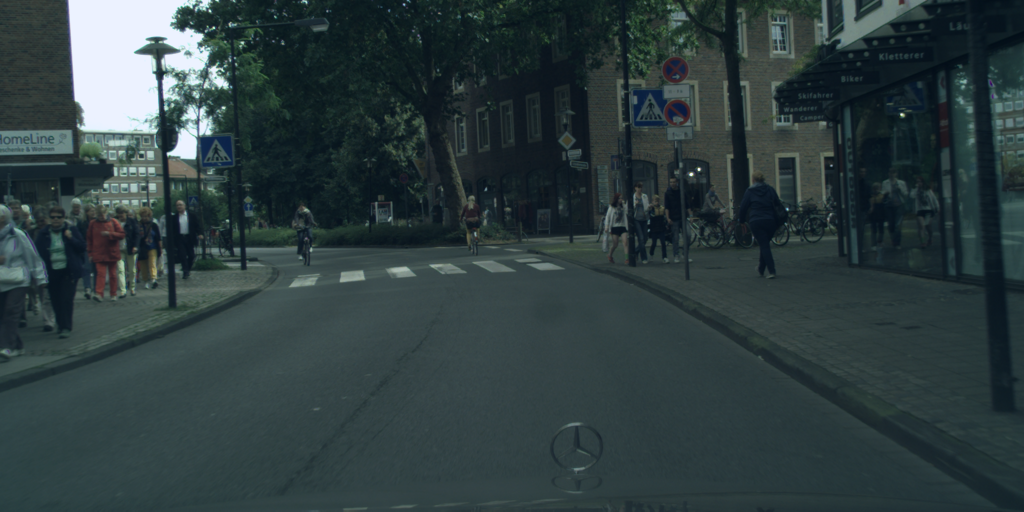} &
   \includegraphics[width=\imgsize]{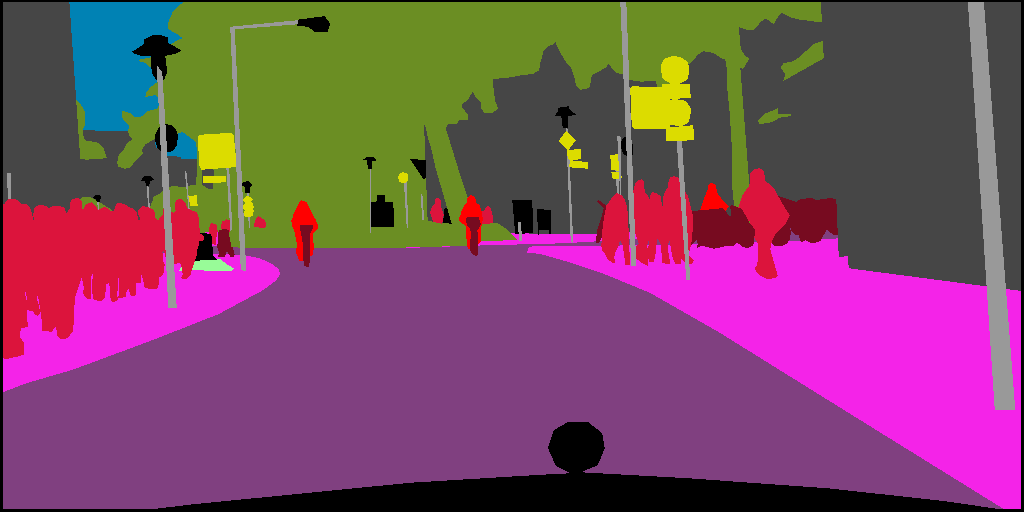} & 
   \includegraphics[width=\imgsize]{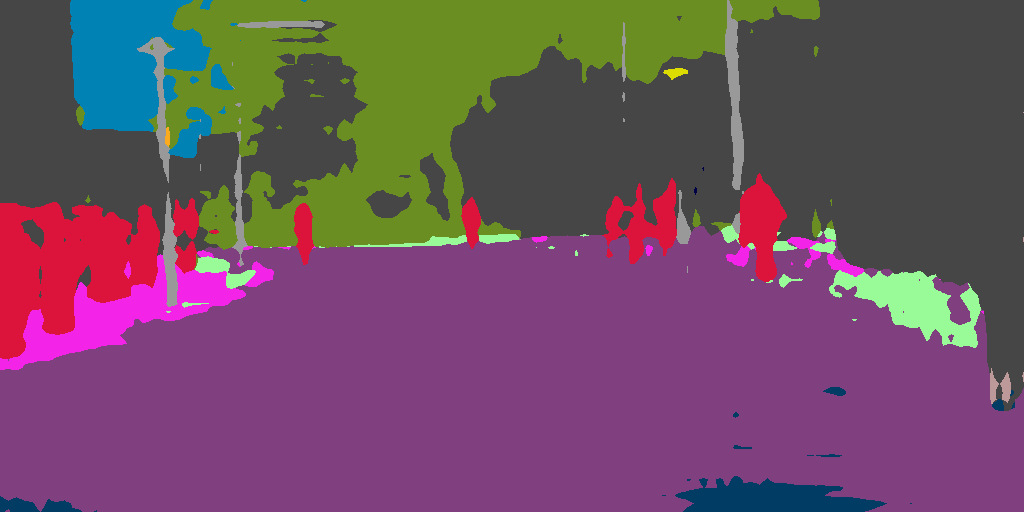} & 
   \includegraphics[width=\imgsize]{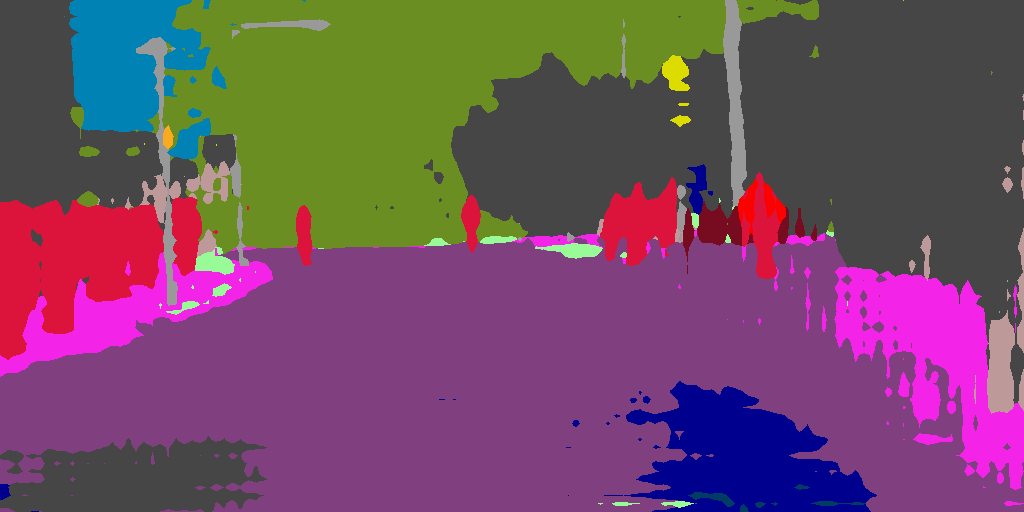} & 
   \includegraphics[width=\imgsize]{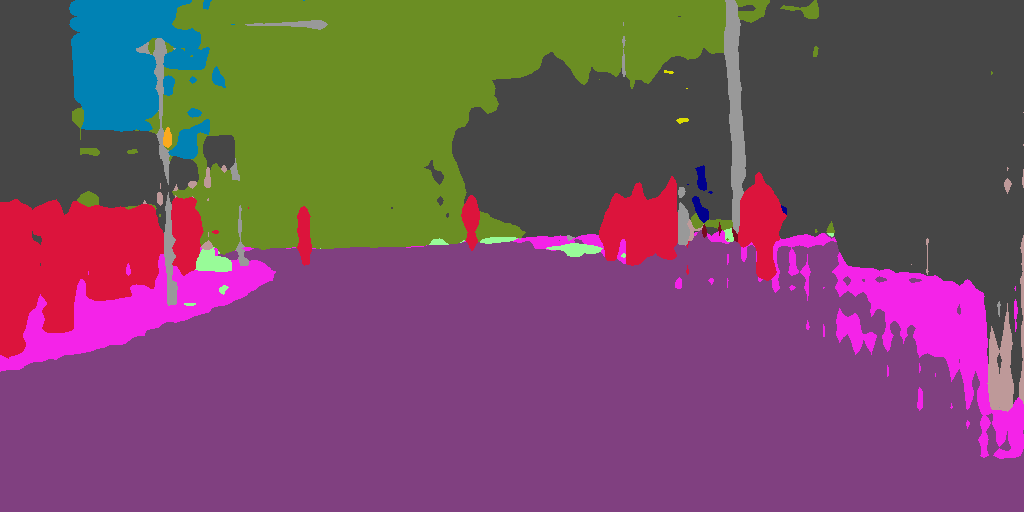} &
   \includegraphics[width=\imgsize]{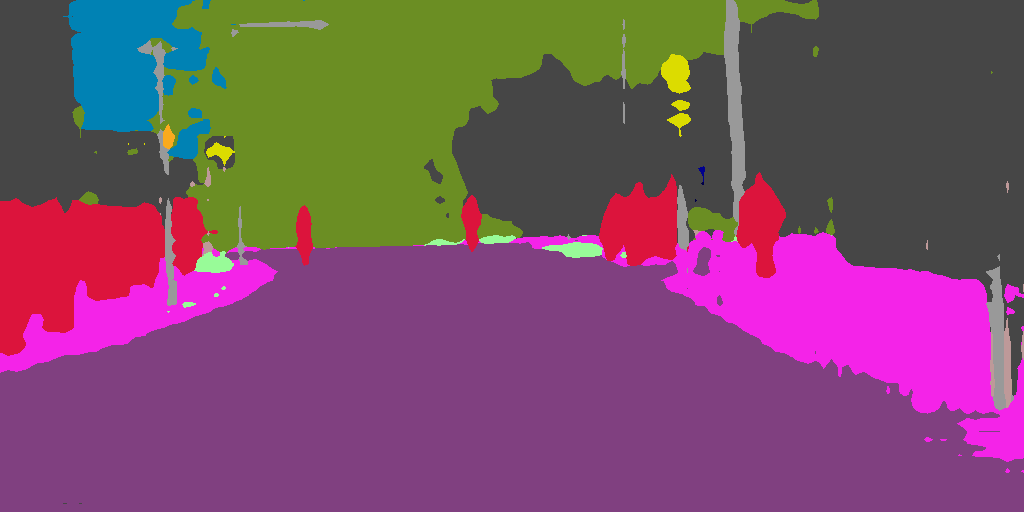} \\
   
   \includegraphics[width=\imgsize]{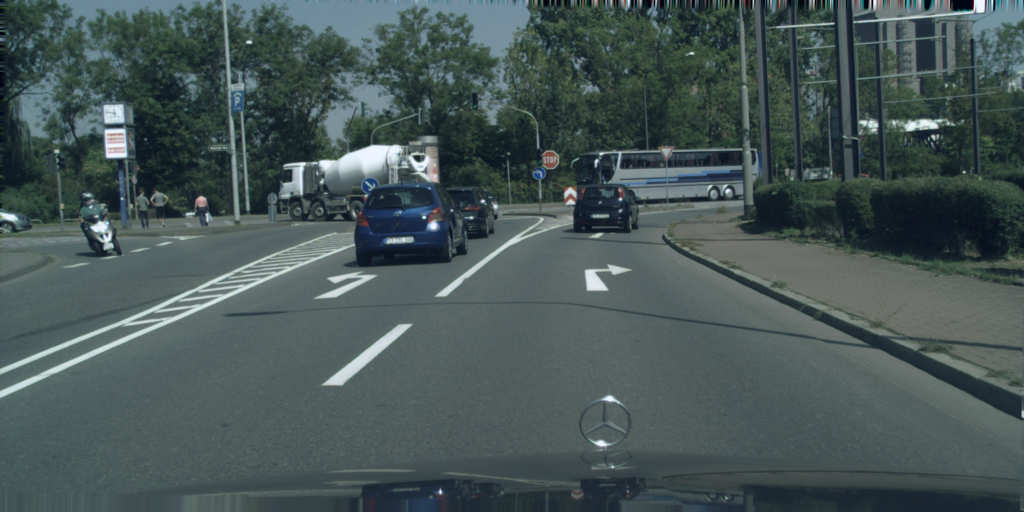} &
   \includegraphics[width=\imgsize]{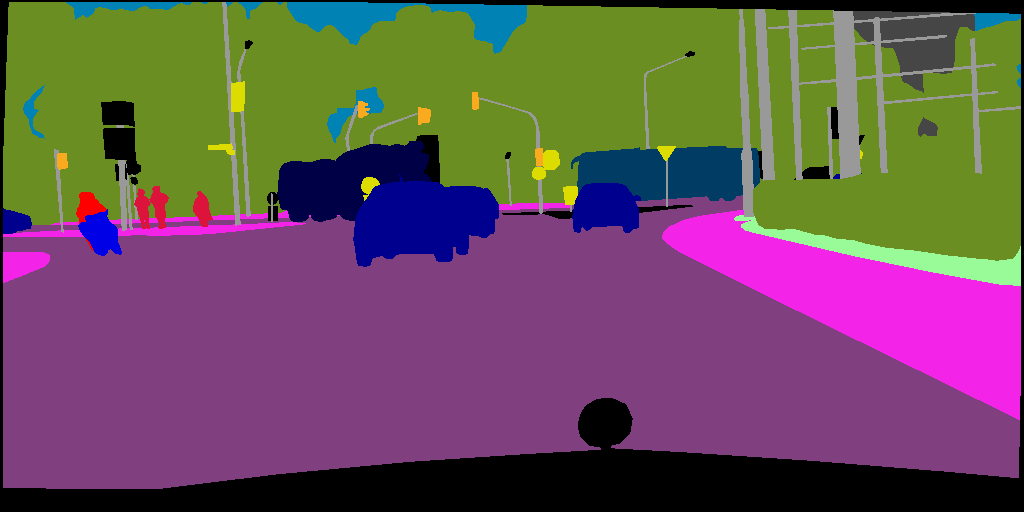} & 
   \includegraphics[width=\imgsize]{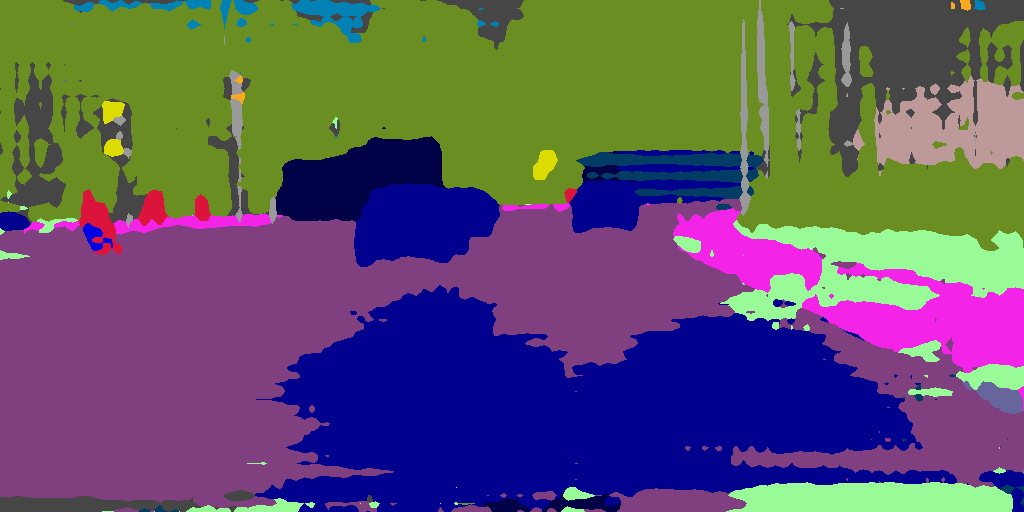} & 
   \includegraphics[width=\imgsize]{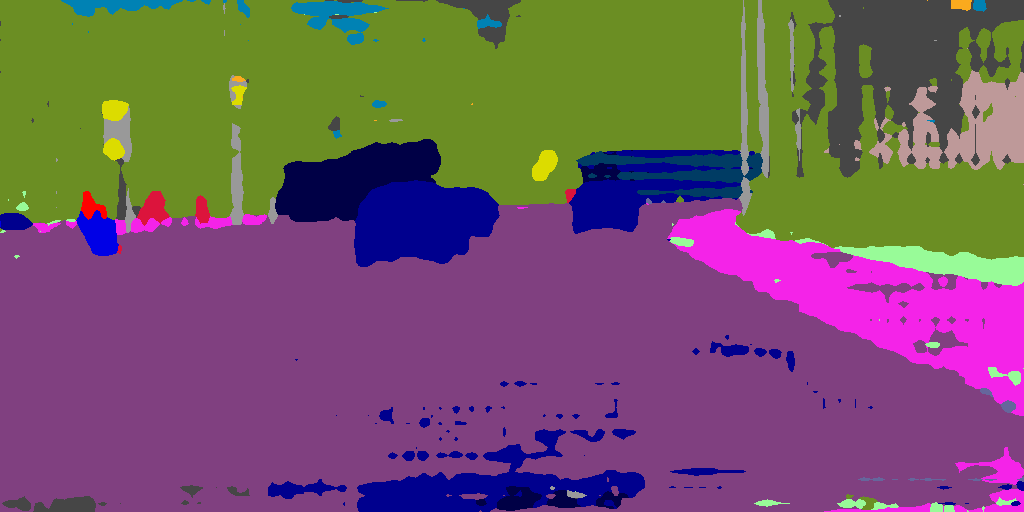} & 
   \includegraphics[width=\imgsize]{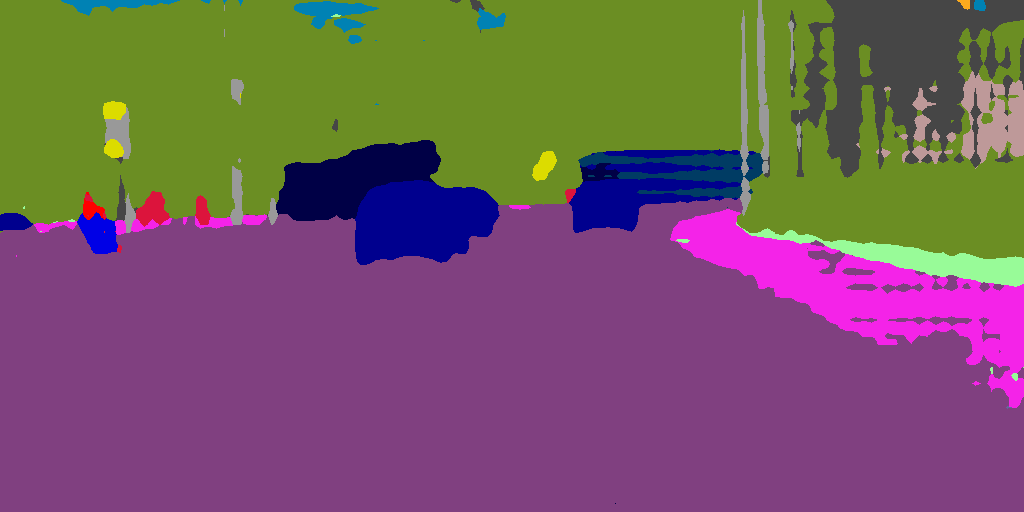} &
   \includegraphics[width=\imgsize]{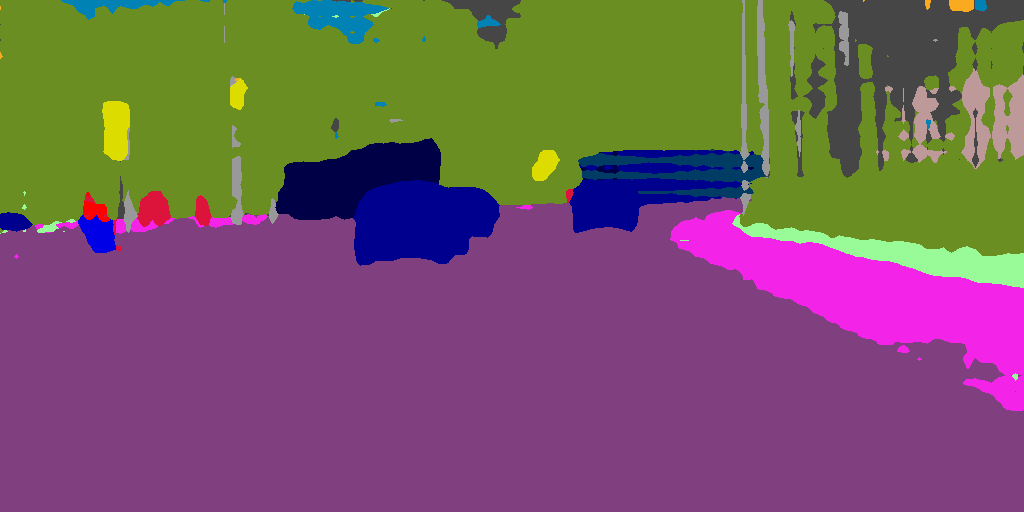} \\ 
   
   Target Images & Ground Truth & Source Only  & Ours ($\mathcal{L}_{cl}$) & MaxSquare IW \cite{Chen2019} & Ours ($\mathcal{L}_{tot}$)

\end{tabular}
\end{subfigure}

\caption{Semantic segmentation of some sample scenes 
from the Cityscapes validation 
dataset when adaptation is performed from the GTA5 source dataset and the DeepLab-V2 with ResNet-101 backbone is employed 
(\textit{best viewed in colors}).}
\label{fig:qual_res}
\end{figure*}

We evaluate the performance of our approach on two widely used synthetic-to-real adaptation scenarios, namely the GTA5 $\rightarrow$ Cityscapes and SYNTHIA $\rightarrow$ Cityscapes benchmarks.
Table \ref{tab:all} reports the numerical results of the experimental evaluation. 
We compare our model to several state-of-the-art methods, which, similarly to our approach, resort to a direct or indirect form of feature-level regularization and distribution alignment to achieve domain adaptation. 
With \textit{source only} we indicate the na\"{i}ve fine-tuning approach, in which no form of target adaptation assists the standard source supervision. 

\subsection{GTA5 $\rightarrow$ Cityscapes}
%
For the GTA5 $\rightarrow$ Cityscapes and ResNet-101 configuration, our approach shows  state-of-the-art performances in feature-level UDA for semantic segmentation, achieving $45.3\%$ of mIoU, which is further boosted up to $45.9\%$ by the entropy minimization objective. 
By looking at Table \ref{tab:all}, we observe about $9\%$ increase over the \textit{source only} baseline, with the improvement well distributed over all the classes. 
A similar behavior can be noted when switching to the less performing VGG-16 backbone: we achieve $33.7\%$ mean IoU with  $\mathcal{L}_{tot}'$ (i.e., without the entropy minimization objective) and $34.2\%$ with $\mathcal{L}_{tot}$ (i.e., with all components enabled) starting from the $25.4\%$ of the baseline  scenario without adaptation. 
\\
When compared with other approaches, our method performs better than standard feature-level adversarial techniques \cite{hoffman2016,hoffman2018,tsai2018,li2020spatial}. 
For example, with the VGG-16 backbone there is a gain of $4.5\%$ w.r.t.\ \cite{hoffman2018}. 
This proves that a more effective class-conditional alignment has been ultimately achieved in the latent space by our approach. 
Due to the similar regularizing effect over feature distribution and comparable ease of implementation, we also compare our framework with some entropy minimization and self-training techniques \cite{vu2019advent,Chen2019,zou2018}, further showing  the effectiveness of our adaptation strategy even if the gap here is a bit more limited. 
With both backbones, our novel feature level modules ($\mathcal{L}_{tot}'$) perform better than \textit{MaxSquare IW} \cite{Chen2019}. 
Moreover, adding the entropy minimization objective from \cite{Chen2019} to $\mathcal{L}_{tot}'$ 
provides a slight but consistent improvement. 
It is worth noting that our method does not rely on additional trainable modules (e.g., adversarial discriminators \cite{hoffman2016,hoffman2018,tsai2018,li2020spatial}) and the whole adaptation process is end-to-end, not requiring multiple separate steps to be re-iterated (e.g., pseudo-labeling in self-training \cite{zou2018}). Moreover, being focused solely on feature level adaptation, it could be easily integrated with other adaptation techniques working at different network levels, such as the input (e.g., generative approaches) or output (e.g., self-training), as shown by the addition of the output-level entropy-minimization loss.

Fig. \ref{fig:qual_res} displays some qualitative results on the Cityscapes validation set of the adaptation process when the ResNet-101 backbone is used. 
%
We observe that the introduction of the clustering module is beneficial to the target segmentation accuracy w.r.t.\ the \textit{source only} case. Some small prediction inaccuracies remain, that are corrected with the introduction of the orthogonality, sparsity and entropy modules in the complete framework. 
By looking at the last two columns, we also notice that our entire framework shows an improvement over the individual entropy-minimization like objective from \cite{Chen2019}, which is reflected in a better detection accuracy both on frequent (e.g.\ \textit{road}, \textit{vegetation}) and less frequent (e.g.\ \textit{traffic sign}, \textit{bus}) classes. 

%
%
%

\subsection{SYNTHIA $\rightarrow$ Cityscapes}
%
To further prove the efficacy of our method, we evaluate it on the more challenging SYNTHIA $\rightarrow$ Cityscapes benchmark, 
where a larger domain gap exists. 
Once more, our approach proves to be successful in performing domain alignment with both ResNet-101 and VGG-16 backbones, reaching state-of-the-art results for feature-level UDA in both configurations (see Table~\ref{tab:all}). 
When ResNet-101 is used, the mIoU\textsuperscript{*} on the 13 classes setting is pushed up to $48.2\%$ from the original $40.5\%$ of \textit{source only}, while the VGG-16 scenario witnesses an even \revised{more} improved performance gain of almost $11\%$  over the no adaptation baseline till a final value of $43.7\%$. 
Differently from the GTA5 $\rightarrow$ Cityscapes case, here the contribution of the entropy-minimization module varies for the two backbones. 
The induced benefit is absent with VGG-16, since the clustering, orthogonality and sparsity jointly enforced  already carry the whole adaptation effort. 
Besides, even the $\mathcal{L}_{em}$ objective alone (i.e., \textit{MaxSquares IW \textsuperscript{(r)}} \cite{Chen2019}) displays quite limited gain over the no adaptation baseline. 
On the contrary, the regularizing effect of the entropy objective is strongly valuable in case the ResNet-101 backbone is used. 
Yet, 
the combination of all modules together actually provides a noticeable boost over both the entropy and feature-level modules separately applied. 
As for the GTA5 scenario, 
our model shows better performance than  feature-level adversarial adaptation \cite{hoffman2016,chen2017nomore,tsai2018,li2020spatial} and output-level approaches \cite{vu2019advent,zou2018} comparable in computational ease. 
Qualitative results of adaptation from SYNTHIA are in the Supplementary Material.

\subsection{Ablation Study}
\label{sec:ablation}

To verify the robustness of the  framework, we perform an extensive ablation study on the adaptation from GTA5 to Cityscapes with ResNet-101 as backbone. First, we examine the contribution of each loss to the final mIoU; then, we  investigate the effect of each novel loss component. Further considerations are  reported in the Supplementary Material.

\begin{table}[tbp]
\setlength{\tabcolsep}{6pt}
 \renewcommand{\arraystretch}{1.1}
\centering
\begin{tabular}{cccc|c}
$\mathcal{L}_{cl}$ & $\mathcal{L}_{or}$ & $\mathcal{L}_{sp}$ & $\mathcal{L}_{em}$ &  mIoU\\[0.1em]
\hline

& & & & 37.0 \\
\checkmark & & & & 42.3 \\
& \checkmark & & & 43.2 \\
& & \checkmark & & 43.7 \\
& & & \checkmark & 44.8 \\
\checkmark & \checkmark & \checkmark &  & 45.3 \\
\checkmark & \checkmark & \checkmark & \checkmark & 45.9 \\

\end{tabular}
\caption{Ablation results on the contribution of each adaptation module in the GTA5 to Cityscapes scenario and with ResNet-101 as backbone.}
\label{tab:ablation}
\end{table}

The contribution of each loss to the adaptation module is shown in Table~\ref{tab:ablation}. Every loss component largely improves the final mIoU results from $37.0\%$ of the \textit{source only} scenario up to a maximum of $44.8\%$.
Combining the $3$ novel modules of this work, we achieve a mIoU of $45.3\%$, which is higher than all the losses alone, but lower than our complete framework with all the losses enabled ($45.9\%$).

\begin{figure}[htbp]
\centering
\includegraphics[trim=2.15cm 0.8cm 2.15cm 0.8cm, clip, width=\linewidth]{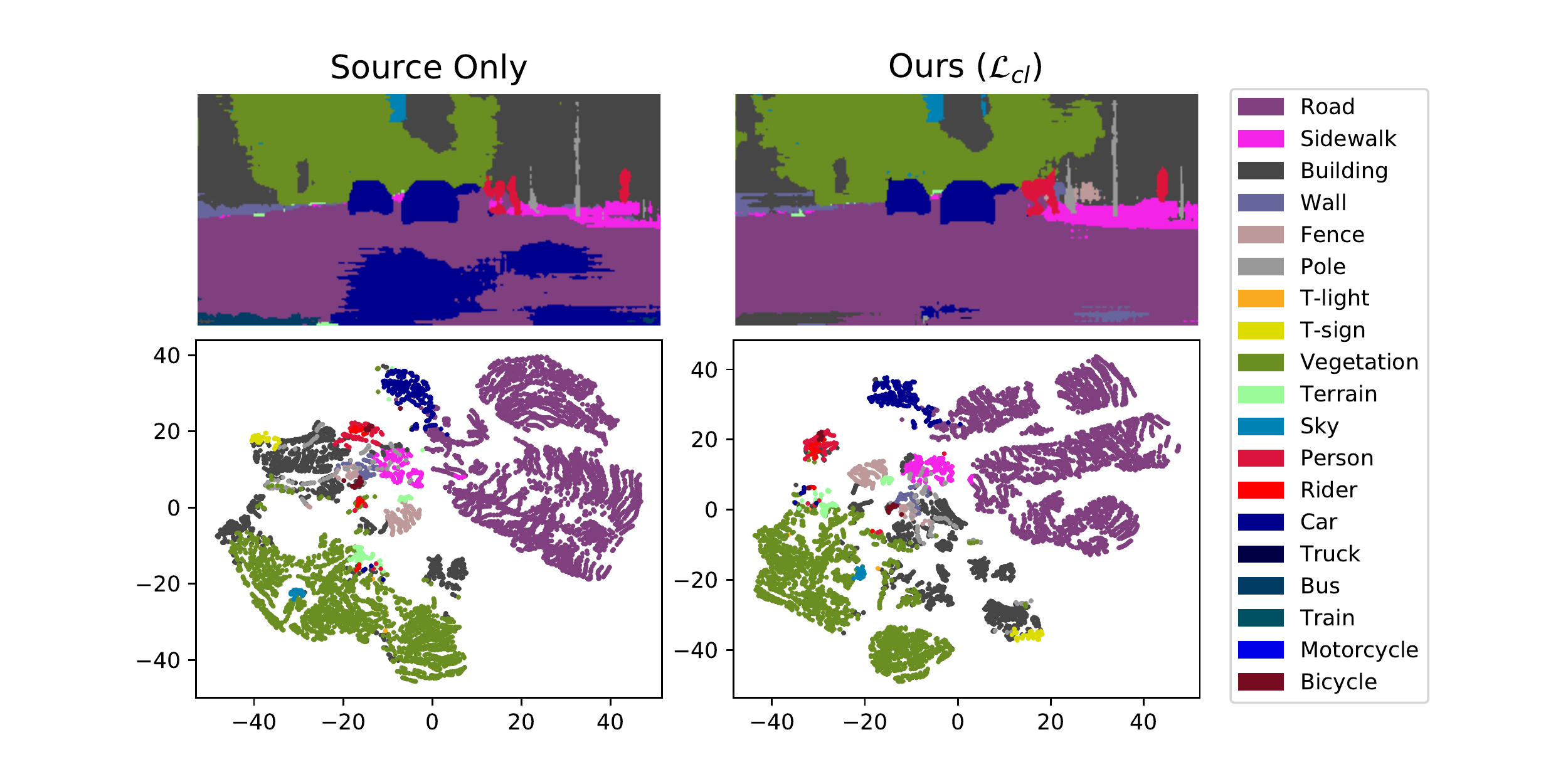}
\caption{T-SNE computed over features of a single image of the Cityscapes validation set when adapting from GTA5 (\textit{best viewed in colors}).}
\label{fig:tSNE}
\end{figure}

To investigate the effect of the clustering module ($\mathcal{L}_{cl}$) we show a t-SNE \cite{maaten2008visualizing} plot of features extracted from a sample image of the Cityscapes validation set. 
In particular, we provide a comparison of the discovered low-dimensional feature distribution for two distinct training settings, i.e.\ \textit{source only} and adaptation with $\mathcal{L}_{cl}$ only. 
The results are reported in Fig.~\ref{fig:tSNE}, where sparse points in the \textit{source only} plot turn out more tightly clustered and spaced apart in the $\mathcal{L}_{cl}$ approach (e.g., look at the \textit{person} class).

\begin{figure}[htbp]
\centering
\includegraphics[trim=0cm 0.5cm 0cm 0.5cm, width=\linewidth]{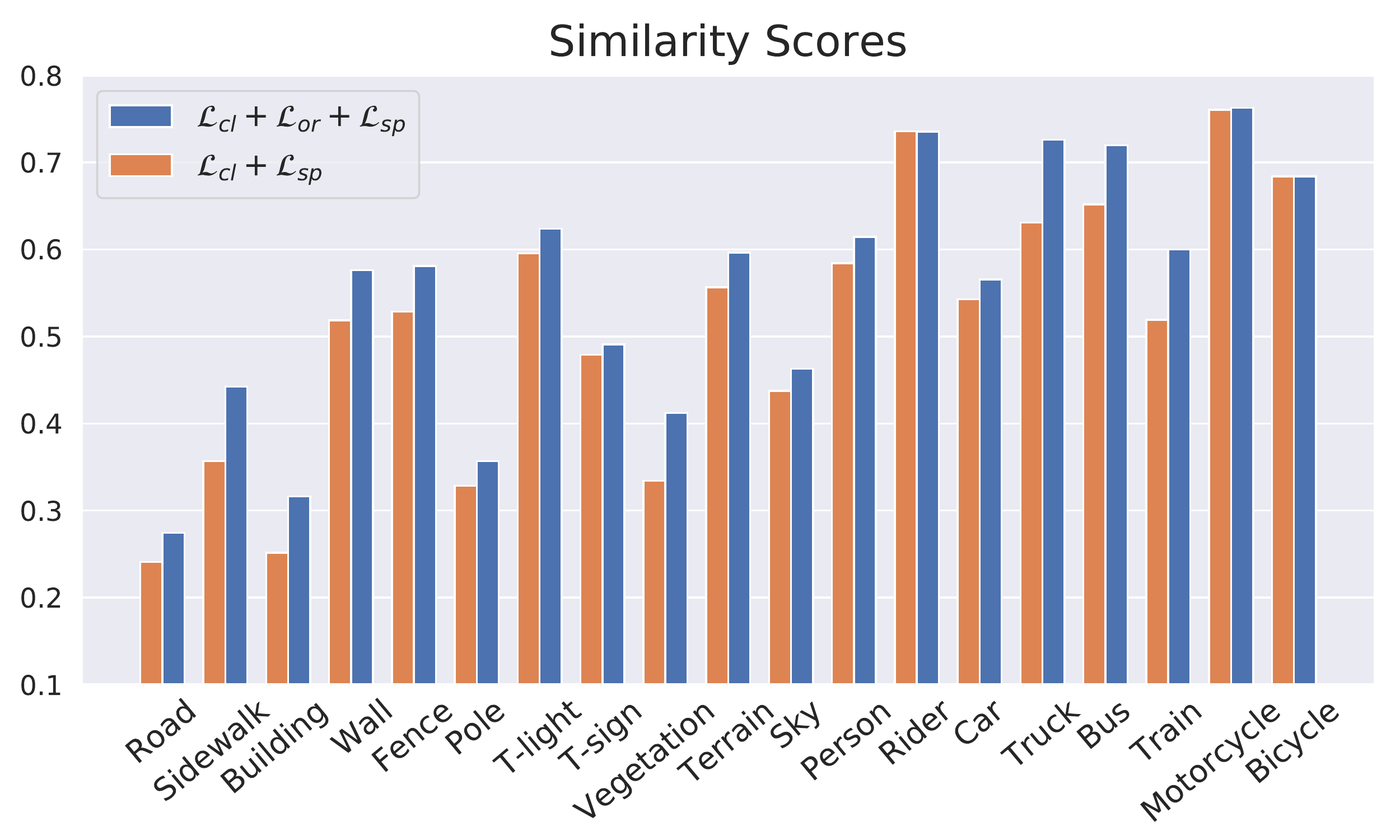}
\caption{Similarity scores computed over images of the Cityscapes validation set when adapting from GTA5.
 }
\label{fig:plot_similarity}
\end{figure}

We analyze the orthogonality constraint ($\mathcal{L}_{or}$) via a similarity score defined as an average class-wise cosine similarity measure (Fig.\ \ref{fig:plot_similarity}). 
The cosine distance is first computed for every pair of feature vectors from a single target image. 
Then, the average values are taken over all features from the same class to get a score for each pair of semantic classes. 
The final values are computed by averaging over all images from the Cityscapes validation set. 
The score computation is performed for 
two different configurations, i.e.\, $\mathcal{L}_{cl}+\mathcal{L}_{or}+\mathcal{L}_{sp}$ and $\mathcal{L}_{cl}+\mathcal{L}_{sp}$, to highlight the effect induced by the orthogonality module. 
Here we report the intra-class similarity scores, whereas the full matrix with also the inter-class values \revised{is} in the Supplementary Material.  
The results in Fig.~\ref{fig:plot_similarity} show that $\mathcal{L}_{or}$ causes the similarity score to significantly increase \revised{within} 
almost all the classes.

%

\begin{figure}[htbp]
\centering
\includegraphics[trim=0cm 0.8cm 0cm 0.8cm, width=\linewidth]{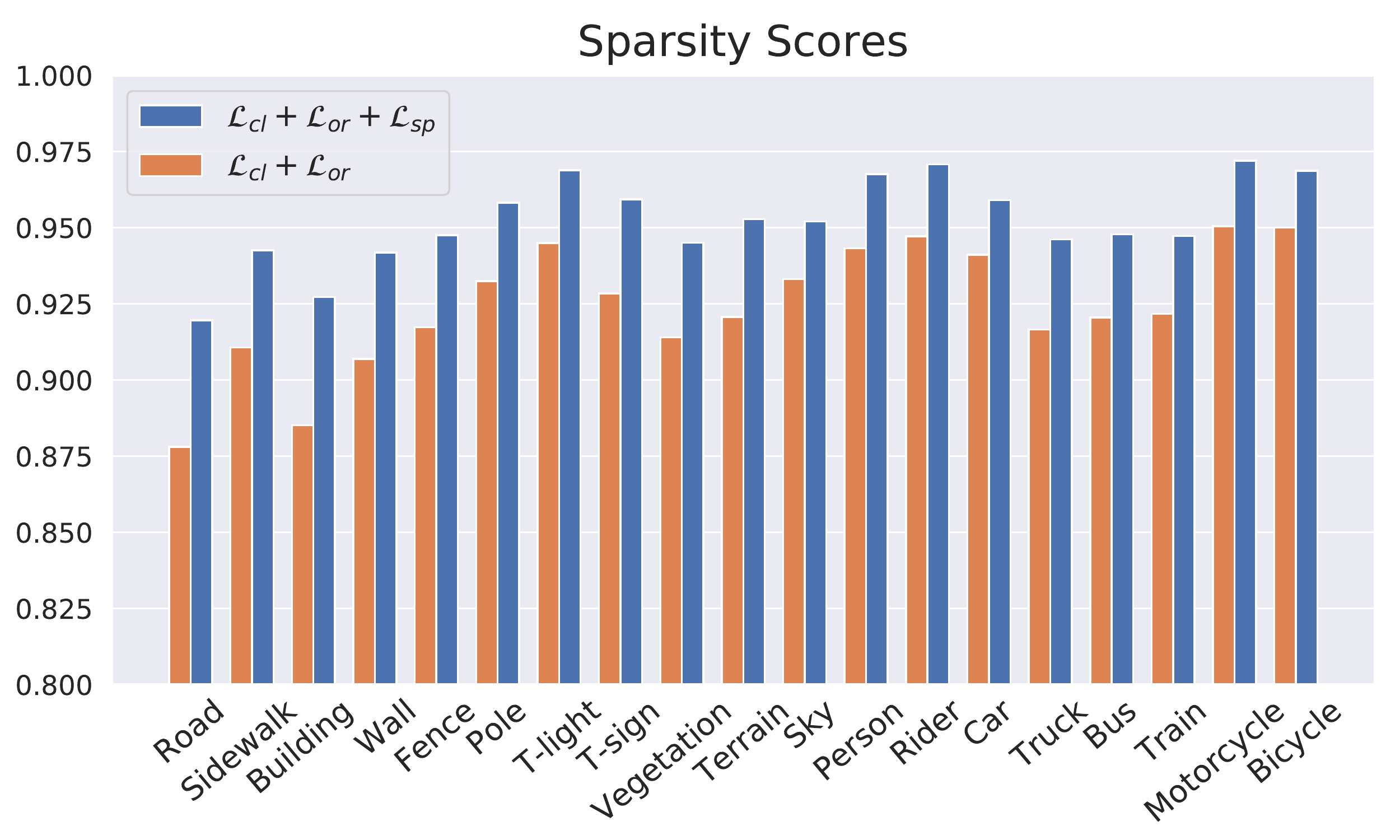}
\caption{Sparsity scores computed over images of the Cityscapes validation set when adapting from GTA5.
}
\label{fig:plot_sparsity}
\end{figure}

To understand the efficacy of the sparsity loss ($\mathcal{L}_{sp}$) we compute the sparsity scores as the fraction of activations in normalized feature vectors close to $0$ or $1$ (Fig. \ref{fig:plot_sparsity}). 
Closeness is quantified as being distant from $0$ or $1$ less than a threshold, which we set to $10^{-4}$. 
As for the similarity scores, the sparsity measures for a single target image are obtained through averaging over all feature vectors from the same class. 
The final results correspond to the mean values over the entire Cityscapes validation set. 
We compute the scores for two different configurations, i.e.\, $\mathcal{L}_{cl}+\mathcal{L}_{or}+\mathcal{L}_{sp}$ and $\mathcal{L}_{cl}+\mathcal{L}_{or}$, so that we can inspect the effect that the sparsity module is providing on feature distribution. 
From Fig.~\ref{fig:plot_sparsity} we can appreciate that $\mathcal{L}_{sp}$ effectively achieves higher values of sparseness for all the classes.

%

\section{Conclusions}



In this paper we propose a novel feature oriented UDA framework for semantic segmentation. 
Our approach comprises 3 main objectives. 
First, features of same class and separate domains are clustered together, whilst features of different classes are spaced apart. 
Second, an orthogonality requirement over the latent space discourages the overlapping of active channels among feature vectors of different classes. 
Third, a sparsity constraint further reduces feature-wise the number of the active channels. 
All combined, these modules allow to reach a 
regularized disposition of latent embeddings 
, while providing a 
semantically consistent domain alignment over the feature space. 
We extensively evaluated our framework in the synthetic-to-real scenario, achieving state-of-the-art results in feature level UDA. 
For future work, we intend to explore new techniques aiming at the refinement of the pseudo-labeling based classification of feature vectors, and to integrate our feature space adaptation with other approaches targeting different network levels. 
\\
\revised{
Finally, 
 we would like to investigate the regularizing effect of the proposed techniques when applied to the single-domain standard semantic segmentation. 
}

{\small
\bibliographystyle{ieee_fullname}
\bibliography{strings,refs}
}

\includepdf[pages={1},offset=0 -0]{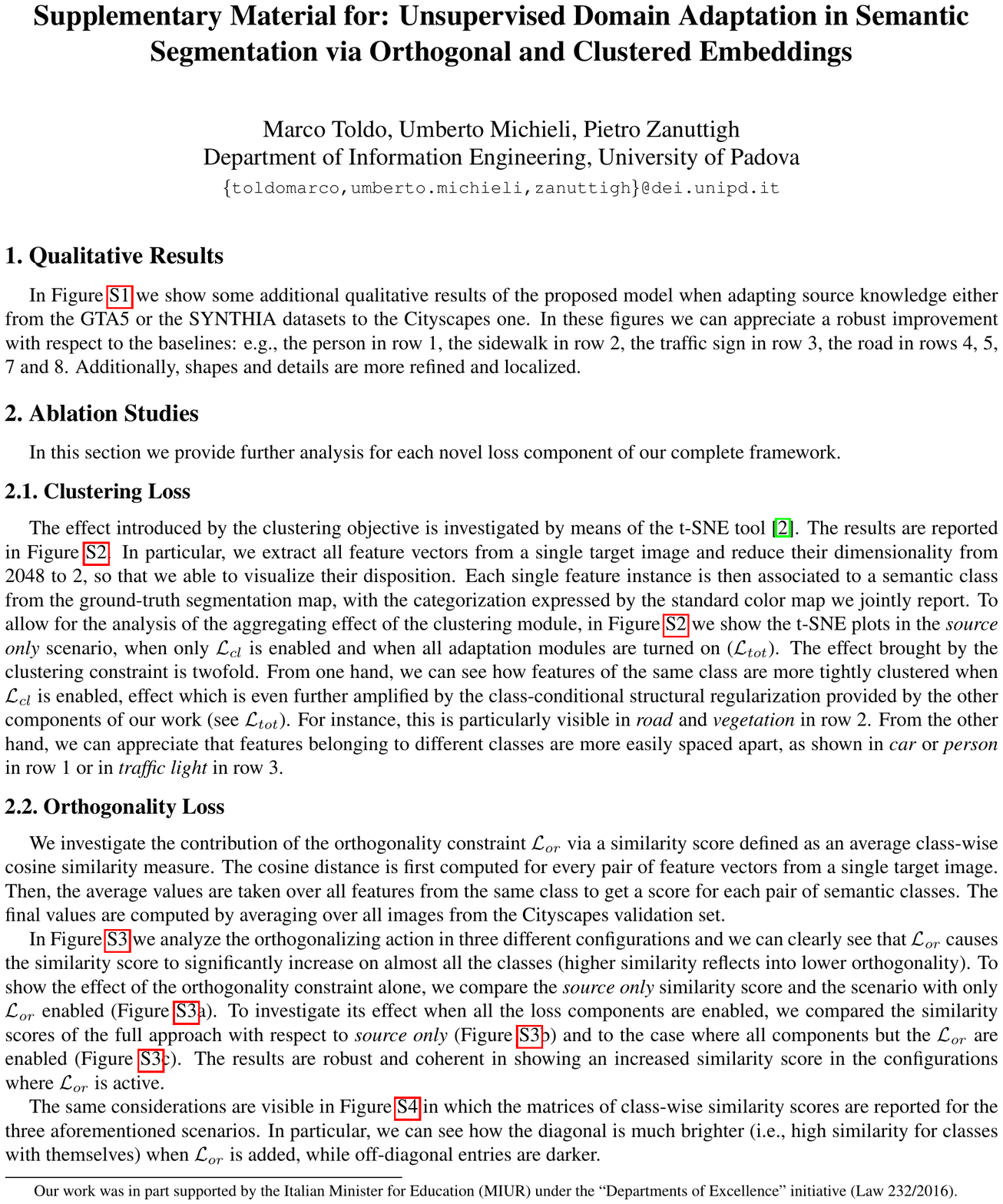}
\includepdf[pages={2},offset=0 -0]{suppl.pdf}
\includepdf[pages={3},offset=0 -0]{suppl.pdf}
\includepdf[pages={4},offset=0 -0]{suppl.pdf}
\includepdf[pages={5},offset=0 -0]{suppl.pdf}
\includepdf[pages={6},offset=0 -0]{suppl.pdf}

\end{document}